\documentclass[letterpaper, 10 pt, conference]{ieeeconf}  

\IEEEoverridecommandlockouts                              

\usepackage{graphics} 
\usepackage{epsfig} 
\usepackage{mathptmx} 
\usepackage{times} 
\usepackage{amsmath} 
\usepackage{amssymb}  

\usepackage{amsthm}

\usepackage{bm}
\usepackage{cite}
\usepackage{xcolor,graphicx}
\usepackage{algorithm}
\usepackage{algpseudocode}
\usepackage{algorithmicx}
\usepackage{subcaption}
\usepackage{overpic}
\usepackage{booktabs}
\usepackage[english]{babel}
\usepackage{url}
\usepackage{placeins}  
\usepackage{afterpage}  
\usepackage{booktabs}  
\usepackage{multirow}  

\usepackage{hyperref}
\usepackage[capitalize,nameinlink]{cleveref}

\crefname{equation}{Eq.}{Eqs.}
\Crefname{equation}{Eq.}{Eqs.}
\usepackage{tikz}
\usepackage{xcolor}

\usepackage{etoolbox}

\usepackage{balance}

\hypersetup{
    colorlinks=false,
    urlbordercolor={1 0 0},
    pdfborder={0 0 1}
}

\usepackage[normalem]{ulem}

\newcommand{\highlight}[1]{#1}

\AtBeginEnvironment{algorithm}{%
  \setlength{\textfloatsep}{5pt plus 1pt minus 2pt}
  \setlength{\intextsep}{5pt plus 1pt minus 2pt}
}

\usepackage{tikz}
\newcommand\copyrightnotice[1]{
    \begin{tikzpicture}[remember picture,overlay]
    \node[anchor=north,xshift=125,yshift=-12pt] at (current page.north) {\parbox{\dimexpr0.45\textwidth-\fboxsep-\fboxrule\relax}{\scriptsize #1}};
    \end{tikzpicture}
}

\title{\LARGE \bf
Multimodal MPC via Bayesian Set Optimization for Dynamic Manipulation of Deformable Linear Objects
}

\author{Youyuan Long$^{1}$, Gokhan Solak$^{1}$, and Arash Ajoudani$^{1}$   
\thanks{This work was supported by the Horizon Europe Project TORNADO (GA 101189557).
}
\thanks{$^{1}$Human-Robot Interfaces and Interaction Lab, Istituto Italiano di Tecnologia, Genova, Italy
        {(e-mail: youyuan.long@iit.it; gokhan.solak@iit.it; arash.ajoudani@iit.it).}}%
}

\begin{document}

\maketitle
\thispagestyle{empty}
\pagestyle{empty}

\begin{abstract}

Dynamic manipulation of deformable linear objects (DLOs) remains challenging due to underactuated, high-dimensional dynamics, delayed deformation responses, and the presence of multiple feasible motions under obstacle constraints. This paper presents BM2PC, a Bayesian Multimodal MPC framework for agile manipulation of DLOs. Building on the control-as-inference posterior, BM2PC formulates multimodal planning as a set optimization problem that balances control quality and behavioral diversity. An efficient incremental update maintains multiple promising control modes with low additional computational overhead. For real-world deployment, BM2PC incorporates an execution dynamics model that predicts actual end-effector motion and compensates for execution mismatch during fast manipulation.
We evaluate BM2PC on rope tracking and obstacle avoidance in simulation and hardware, and on simulated disturbance rejection. In trajectory tracking, BM2PC generalizes better to unseen trajectories than a learning-based baseline without task-specific policy training or fine-tuning. Real-world experiments demonstrate accurate long-horizon tracking, reducing RMSE by 44.0\% on the longest task. In obstacle avoidance, BM2PC achieves comparable performance to existing multimodal MPC methods while reducing planning time by 29\% to 66\%. 
These results demonstrate accurate dynamic control and efficient multimodal planning within a unified MPC framework. 
The core implementation of BM2PC and supplementary videos are available at
\url{https://github.com/Longyouyuan/BM2PC}.
\end{abstract}

\copyrightnotice{\copyright 2026 IEEE. 
This work has been submitted to the IEEE for possible publication. Copyright may be transferred without notice, after which this version may no longer be accessible.}

\vspace{0.5em}

\begingroup
\small
\bfseries
\noindent\hspace{1em}%
\textit{Note to Practitioners---}%
This work is motivated by industrial tasks in which robots must move ropes, cables, and hoses, such as cable routing, obstacle-aware transportation, and handling flexible components in crowded workspaces.
Existing solutions either require training a new controller for each task or, in the case of optimization-based methods, incur considerable computation to maintain multiple alternative motion strategies.
To address this limitation, the proposed framework preserves several promising control solutions through a lightweight update, allowing the robot to adapt its motion when obstacles move, disturbances occur, or the preferred solution changes.
An important practical component of our method is the execution-dynamics model, which compensates for delays and tracking errors that are negligible in simulation but can prevent reliable high-speed manipulation on physical hardware.
Experiments demonstrate accurate tracking and obstacle avoidance in simulation and hardware, disturbance recovery in simulation, and faster planning than representative multimodal methods, without task-specific training.
The current implementation assumes that the rope state and obstacle geometry are provided by an external perception system and has been tested with a rope grasped at one end. Future deployment will require more robust onboard perception, automatic adaptation to different cables and hoses, and validation in other potential applications, such as automated cable installation, flexible-part assembly, and tether management.
\par
\endgroup

\vspace{0.5em}

\begingroup
\small
\bfseries
\noindent\hspace{1em}%
\par
\endgroup
\section{INTRODUCTION}
Robotic manipulation of deformable linear objects (DLOs), such as ropes, cables, and hoses, is an important capability for service, industrial, and field robots~\cite{caporali2026robotic, sanchez2018robotic}.
In applications such as automated cable routing, wire-harness assembly, and hose handling, robots must manipulate DLOs accurately and efficiently through structured workspaces containing fixtures and obstacles.
Compared with rigid-body manipulation, DLO manipulation is substantially more challenging due to its high-dimensional, underactuated, and highly nonlinear dynamics~\cite{caporali2026robotic, yin2021modeling, gu2023survey}.
These challenges become more pronounced in dynamic manipulation, where fast motions induce large deformations and complex transient behaviors~\cite{chi2024iterative}.
In environments involving obstacles, disturbances, and changing task conditions, multiple feasible motion strategies, referred to as control modes in this work, may coexist, with their relative desirability changing over time.
Consequently, practical control requires accurately handling complex deformable dynamics while preserving alternative motion strategies that can be rapidly selected when the current strategy becomes infeasible.

Previous advances in dynamic DLO manipulation can be broadly grouped into two categories: \textit{learning-based} and \textit{optimization-based} control approaches. 
Learning-based methods shift the computational burden to an offline training phase by learning control policies~\cite{long2026self,chen2023dynamic}, enabling fast execution but typically requiring extensive training data and exhibiting limited generalization to unseen trajectories, object properties, and changing environments.
In contrast, optimization-based methods compute control actions online using predictive models~\cite{nah2021manipulating,gabellieri2025manipulation}, naturally generalizing across tasks without retraining and readily accommodating changing objectives and constraints.
To handle environments with multiple feasible solutions, several multimodal optimization methods have been proposed~\cite{park2025csc,lambert2020stein}. 
While these approaches improve planning robustness by maintaining diverse candidate solutions, they introduce additional computational overhead that limits their applicability to agile real-time manipulation.
Consequently, there remains a need for optimization-based methods that can efficiently maintain multiple feasible control strategies while preserving the computational efficiency required for real-time dynamic DLO manipulation.

To address this gap, we propose BM2PC, a sampling-based MPC framework for efficient multimodal planning without task-specific policy training. Building on the control-as-inference posterior, BM2PC formulates multimodal planning as a set optimization problem that jointly accounts for the posterior quality and behavioral diversity of control sequences. Rather than approximating the full posterior distribution, the method directly selects a compact set of promising control hypotheses through an efficient incremental update. The selected modes are propagated across MPC steps, allowing the controller to preserve alternative solutions and adapt to changing constraints with low computational overhead.

We evaluate BM2PC on dynamic rope manipulation, where the robot controls the grasped endpoint to track a desired rope-tip trajectory. Our evaluation focuses on two complementary capabilities. First, complex trajectory tracking and disturbance rejection demonstrate accurate predictive control of highly nonlinear rope dynamics with delayed responses. Compared with a learning-based method, BM2PC shows stronger generalization to unseen trajectories without task-specific policy training or fine-tuning. Second, obstacle avoidance evaluates multimodal planning under changing constraints. Compared with existing multimodal MPC methods, BM2PC achieves comparable control performance while reducing planning time by 29\% to 66\%. Further ablations demonstrate the benefits of maintaining diverse control modes when obstacles suddenly change, while comparison with exhaustive optimization confirms that the incremental update achieves high solution quality at substantially lower computational cost.


Real-world deployment introduces an additional source of prediction error, as commanded end-effector (EE) motions may deviate from their actual execution due to communication delays, actuator limits, and low-level tracking errors. To compensate for these effects, we incorporate an \textit{execution dynamics model} that predicts the realized EE motion from commanded trajectories and recent execution history. This execution-aware prediction improves planning accuracy during fast dynamic manipulation and enables reliable real-world deployment of BM2PC.

The main contributions of this paper are summarized as follows:
\begin{itemize}
    \item We formulate multimodal MPC as Bayesian set optimization over finite-horizon control sequences and develop an efficient incremental approximation that maintains high-quality, behaviorally diverse control modes for real-time planning.

    \item We enable reliable real-world deployment of BM2PC through an execution-aware predictive model that compensates for execution mismatch by incorporating predicted EE motion into the planning process.

    \item We validate BM2PC through extensive simulation and real-world experiments, demonstrating accurate dynamic manipulation of unseen trajectories without task-specific policy training and efficient multimodal planning under changing constraints.
\end{itemize}

\section{Related work}

\subsection{Manipulation of Deformable Linear Objects}
DLO manipulation has been studied across a wide range of tasks, including knot tying, cable routing, insertion, shape control, transportation, and obstacle avoidance~\cite{jqr-46-5-623, saha2006motion, sudry2023hierarchical, aksoy2025planning, monguzzi2025optimal, almaghout2024robotic, ying2024obstacle, tang2018framework}. Classical planning approaches often exploit geometric or topological abstractions. For instance, early work used topologically biased planning for knotting with dual-arm robots~\cite{saha2006motion}, while later studies combined knot theory, learned inverse models, and hierarchical planning for rope manipulation~\cite{sudry2023hierarchical}. In industrially motivated cable and wire-harness scenarios, model-based planning and control have been used for routing, shape regulation, and collision-free transportation~\cite{aksoy2025planning, monguzzi2025optimal, almaghout2024robotic, ying2024obstacle, tang2018framework}. These methods are effective for structured tasks, but many of them assume quasi-static or slowly varying deformation, where inertial effects and phase lag are limited. In contrast, agile DLO manipulation requires predicting delayed and oscillatory responses caused by fast endpoint motion, which substantially increases the difficulty of online control.

Recent efforts have increasingly focused on dynamic manipulation of DLOs, where inertial effects, oscillation, and delayed motion propagation become dominant. 
Existing methods can be broadly grouped into \textit{optimization-based} and \textit{learning-based} approaches. 

Optimization-based methods search for motion commands or trajectory parameters that induce desired dynamic behaviors. 
Some works reduce the search space through parameterized motions or dynamic primitives, such as 3D whip manipulation with low-dimensional primitives~\cite{nah2021manipulating} and Iterative Residual Policy (IRP) for goal-conditioned dynamic manipulation~\cite{chi2024iterative}. 
While effective for repeatable behaviors, such methods often depend on carefully designed motion parameterizations or iterative task-level refinement, limiting flexibility under changing task geometry, constraints, or objectives. 
Other works embed DLO dynamics into online optimal control, including cable shape control with reduced-order spectral models~\cite{shen2025aerial}, dynamically feasible trajectory generation for UAV--cable systems~\cite{gabellieri2025manipulation}, and tethered-aerial-vehicle NMPC under cable-induced coupling constraints~\cite{prkavcin2025trajectory}. 
These methods provide explicit predictive reasoning and constraint handling, but their iterative numerical optimization can become expensive as model complexity, horizon length, and environmental constraints increase, especially when complex neural components such as learned execution dynamics are included.

Learning-based methods provide an alternative by amortizing control computation into trained policies or generative models. 
SPiD combines physics-informed self-supervised learning with a mass-spring model extended with dynamic parameters~\cite{long2026self}; simulation-trained policies have been transferred to hardware for dynamic wire flinging using discrete elastic rods~\cite{chen2023dynamic}; and Dynamics-Informed Diffusion Policy has been proposed for 3D goal-conditioned rope manipulation with physics-informed test-time adaptation~\cite{lan2025dynamic}. 
Reinforcement learning approaches have also been explored for goal-conditioned dynamic DLO manipulation and oscillation avoidance in cable manipulation~\cite{zhaole2024dexdlo, liuni2026dyndlo}.
Although learning-based controllers can achieve fast execution after training, they typically require task-specific data, demonstrations, or extensive simulation training, and their generalization to arbitrary trajectories, dynamic obstacles, and changing execution conditions remains challenging.

In contrast, BM2PC retains the flexibility of online sampling-based MPC while avoiding expensive iterative trajectory optimization. 
By evaluating candidate control sequences through parallel rollouts, it naturally supports non-smooth costs, collision checking, and complex predictive models such as the proposed execution dynamics model. 
Its almost-free multimodal update further enables real-time dynamic DLO manipulation without expert demonstrations or task-specific policy training.

\subsection{Sampling-Based and Multimodal Model Predictive Control}
Model predictive control is attractive for dynamic DLO manipulation because it replans online from the current state and can directly incorporate nonlinear dynamics, task costs, and collision constraints. For complex systems where gradients are unreliable or costs are non-smooth, sampling-based MPC is particularly useful. Model Predictive Path Integral control (MPPI) and information-theoretic MPC evaluate many stochastic rollouts in parallel and update control sequences through cost-weighted sampling~\cite{williams2017model, williams2018information}. These methods are well suited to GPU implementation and have been widely used for nonlinear robotic systems. However, standard MPPI typically maintains a single nominal control sequence or a unimodal sampling distribution, which can be problematic when the optimal-control landscape contains multiple separated low-cost regions~\cite{toda2021clustering}.

Several methods have attempted to improve sampling-based MPC in multimodal or constrained settings. One direction improves the proposal distribution: learned normalizing-flow samplers can generate more informative control samples~\cite{sacks2023learning}, and Biased-MPPI allows arbitrary sampling distributions informed by ancillary controllers to reduce local-minimum sensitivity~\cite{trevisan2024biased}. Another direction separates rollout modes through clustering. Rollout-clustering MPPI shows that averaging samples from different modes can generate unsafe controls and uses trajectory clustering to avoid such failures~\cite{patrick2024path}. CSC-MPPI further combines DBSCAN clustering with constrained MPPI for reliable obstacle avoidance~\cite{park2025csc}. Bayesian particle-based formulations provide another route. SVMPC formulates MPC as posterior inference over control parameters and uses Stein variational gradient descent to represent multimodal solution distributions~\cite{lambert2020stein}, while SVG-MPPI introduces Stein variational guidance to handle rapidly changing multimodal action distributions~\cite{honda2024stein}. 

These studies have extensively investigated the limitations of single-mode sampling in multimodal control landscapes and demonstrated the benefit of maintaining diverse solutions. However, existing multimodal approaches often incur additional computational costs through clustering, density estimation, gradient-based particle transport, or iterative distribution updates. Rather than revisiting the general comparison between single-mode and multimodal MPC, this work focuses on developing a computationally efficient multimodal MPC framework. BM2PC achieves this through a lightweight incremental multimodal update over sampled rollouts and is evaluated against representative multimodal MPC methods under matched planning conditions.

\section{Preliminary}

\subsection{Finite-horizon optimal control}

At each MPC step, we consider a finite-horizon optimal control problem starting from the currently observed state $\bm{x}_0$.
For notational simplicity, the horizon-local indices are reset to $t=0,\dots,T-1$.
Let $\bm{x}_t\in\mathbb{R}^{n_x}$ and $\bm{u}_t\in\mathbb{R}^{n_u}$ denote the state and control input at stage $t$, respectively.
We use the shorthand $\bm{u}_{0:T-1}:=[\bm{u}_0,\dots,\bm{u}_{T-1}]$ and $\bm{x}_{0:T}:=[\bm{x}_0,\dots,\bm{x}_T]$.

The system dynamics are described by $\bm{x}_{t+1}=f(\bm{x}_t,\bm{u}_t)$.
Given the current state $\bm{x}_0$ and a control sequence $\bm{u}_{0:T-1}$, the resulting state trajectory is denoted by $\bm{x}_{0:T}(\bm{u}_{0:T-1})$.

We define the finite-horizon trajectory cost as
\[
J(\bm{x}_{0:T},\bm{u}_{0:T-1})
=
c_T(\bm{x}_T)
+
\sum_{t=0}^{T-1}
c(\bm{x}_t,\bm{u}_t),
\]
where $c(\cdot,\cdot)$ is the stage cost and $c_T(\cdot)$ is the terminal cost.

The corresponding rollout cost of a control sequence is
\[
J_T(\bm{u}_{0:T-1})
:=
J\!\left(
\bm{x}_{0:T}(\bm{u}_{0:T-1}),
\bm{u}_{0:T-1}
\right).
\]

The finite-horizon optimal control problem is therefore
\[
\min_{\bm{u}_{0:T-1}}
\; J_T(\bm{u}_{0:T-1}).
\]

\subsection{Control as probabilistic inference}
Instead of viewing MPC solely as deterministic optimization, we adopt the control-as-inference perspective \cite{honda2025model} and introduce an auxiliary binary optimality variable $\mathcal{O}_{\tau}\in\{0,1\}$, where $\mathcal{O}_{\tau}=1$ indicates that the trajectory $\tau:=[\bm{x}_{0:T},\bm{u}_{0:T-1}]$ is optimal.
For brevity, we denote the event $\mathcal{O}_{\tau}=1$ simply as $\mathcal{O}_{\tau}$ in the following.

Given $\bm{x}_0$ and $\bm{u}_{0:T-1}$, the induced trajectory and its rollout cost are uniquely determined.
Following the standard exponentiated-utility construction, we define the optimality likelihood as
\[
p(\mathcal{O}_{\tau}\mid\bm{u}_{0:T-1},\bm{x}_0)
\propto
\exp\!\bigl(-\lambda^{-1}J_T(\bm{u}_{0:T-1})\bigr),
\]
where $\lambda>0$ is a temperature parameter.

Conditioned on the current state $\bm{x}_0$, Bayes' rule gives the posterior
\[
p(\bm{u}_{0:T-1}\mid\mathcal{O}_{\tau},\bm{x}_0)
=
\frac{
p(\mathcal{O}_{\tau}\mid\bm{u}_{0:T-1},\bm{x}_0)
\,p(\bm{u}_{0:T-1})
}{
\int
p(\mathcal{O}_{\tau}\mid\tilde{\bm{u}}_{0:T-1},\bm{x}_0)
\,p(\tilde{\bm{u}}_{0:T-1})
\,d\tilde{\bm{u}}_{0:T-1}
},
\]
where $p(\bm{u}_{0:T-1})$ denotes a prior distribution over control sequences.

We introduce a candidate distribution $\pi(\bm{u}_{0:T-1})$ over admissible control sequences and define the optimal variational approximation as
\[
\pi^\ast
=
\arg\min_{\pi\in\mathcal{P}}
D_{\mathrm{KL}}
\!\left(
\pi(\bm{u}_{0:T-1})
\;\middle\|\;
p(\bm{u}_{0:T-1}\mid\mathcal{O}_{\tau},\bm{x}_0)
\right),
\]
where $\mathcal{P}$ denotes the chosen family of distributions over control sequences.
Using Bayes' rule together with the optimality likelihood above, this problem can be equivalently written as
\begin{equation*}
\begin{aligned}
\pi^\ast
=
\arg\min_{\pi\in\mathcal{P}}
\Bigl\{
&\lambda^{-1}
\mathbb{E}_{\pi}
\!\left[
J_T(\bm{u}_{0:T-1})
\right] \\
&\quad+
D_{\mathrm{KL}}
\!\left(
\pi(\bm{u}_{0:T-1})
\|p(\bm{u}_{0:T-1})
\right)
\Bigr\}.
\end{aligned}
\end{equation*}

When $\mathcal{P}$ contains all probability densities, the variational problem recovers the optimal control distribution
\begin{equation}
\pi^\ast(\bm{u}_{0:T-1})
=
Z^{-1}
\exp\!\bigl(-\lambda^{-1}J_T(\bm{u}_{0:T-1})\bigr)
\,p(\bm{u}_{0:T-1}),
\label{pi star}
\end{equation}
where
\[
Z
=
\int
\exp\!\bigl(-\lambda^{-1}J_T(\bm{u}_{0:T-1})\bigr)
\,p(\bm{u}_{0:T-1})
\,d\bm{u}_{0:T-1}.
\]

A detailed discussion of MPC from the probabilistic-inference perspective can be found in \cite{honda2025model}.

\section{Method}
\begin{figure}[h]
    \centering
    \includegraphics[width=.92\linewidth]{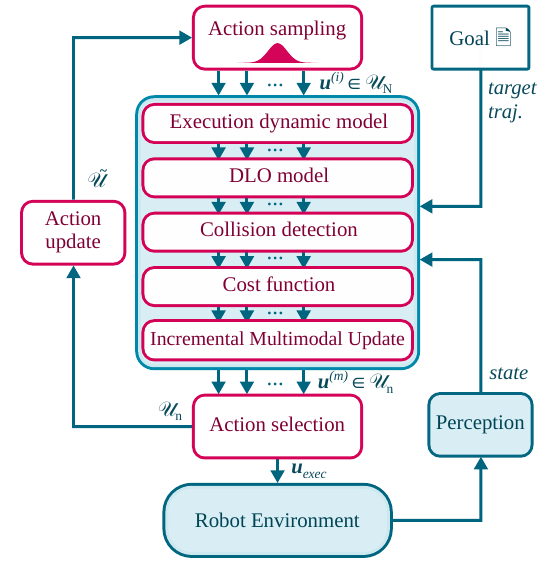}
    \vspace{-0.1cm}
    \caption{\textbf{Overview of the proposed BM2PC framework.}
Candidate control sequences are sampled around the multimodal warm-start set and evaluated through parallel rollouts using the execution dynamics model, DLO dynamics model, and collision detection. Based on rollout costs and trajectory features, BM2PC performs an incremental multimodal update to maintain a diverse set of high-quality control modes. One mode is selected for execution, while the updated multimodal set is propagated to the next MPC iteration.}
    \label{fig:method-overview}
    \vspace{-0.4cm}
\end{figure}

In this section, we describe the complete BM2PC framework shown in Fig.~\ref{fig:method-overview}. The proposed approach combines multimodal planning, predictive DLO simulation, and execution-aware motion prediction within a unified MPC pipeline. A key component is the incremental multimodal update, which efficiently constructs multiple diverse low-cost control hypotheses during optimization. Candidate trajectories are evaluated using both the DLO dynamics model and the execution dynamics model, allowing the planner to account for complex rope dynamic responses as well as real-world command-execution mismatch. The following subsections describe BM2PC, the DLO model, the execution dynamics model, and collision detection in detail.

\subsection{BM2PC}
\subsubsection{Problem formulation}
Instead of seeking a single optimal control sequence, BM2PC aims to identify a finite set of control sequences that have \textbf{high probability under the optimal control distribution} while maintaining \textbf{sufficient separation} in trajectory-feature space, thereby preserving multiple promising modes simultaneously.
Let
\[
\mathcal U
=
\left\{
\bm{u}^{(k)}_{0:T-1}
\right\}_{k=1}^{n}
\]
denote a set of \(n\) control sequences. We formulate the problem as
\begin{equation}
\begin{aligned}
\max_{\mathcal U}
&\;
\prod_{k=1}^{n}
\pi^\ast\!\left(\bm{u}^{(k)}_{0:T-1}\right)\,
\Psi(\mathcal U) \label{BM2PC obj}
\end{aligned}
\end{equation}
where \(\pi^\ast(\bm{u}_{0:T-1})\) is the optimal control distribution in \eqref{pi star}, and \(\Psi(\mathcal U)\) promotes behavioral diversity among the selected control sequences. The former is derived from the standard control-as-inference framework, while the latter explicitly couples the individual control hypotheses at the set level. Consequently, the proposed formulation directly optimizes a finite set of high-posterior, behaviorally diverse control sequences rather than approximating the full posterior distribution. We refer to this posterior-guided formulation as Bayesian set optimization, distinct from surrogate-based Bayesian optimization.

We define the diversity measure as
\begin{equation}
\begin{aligned}
\Psi(\mathcal U)
&=
\exp\!\left(
\beta
\sum_{i<j}
\left\|
\phi\!\left(\tau(\bm{u}^{(i)}_{0:T-1})\right)
-
\phi\!\left(\tau(\bm{u}^{(j)}_{0:T-1})\right)
\right\|_2^2
\right) \label{diversity measure}
\end{aligned}
\end{equation}
where \(\beta>0\) controls the strength of diversity promotion, \(\tau(\bm{u}^{(i)}_{0:T-1})\) denotes the trajectory induced by the control sequence \(\bm{u}^{(i)}_{0:T-1}\), and \(\phi(\cdot)\) is a feature map defined on trajectories. \highlight{In our experiments, $\phi(\tau)$ is a lightweight feature extractor that retains only the predicted rope-tip position trajectory over the planning horizon, enabling efficient comparison of candidate behaviors.}

Substituting \cref{pi star} and \cref{diversity measure} into \cref{BM2PC obj} and simplifying, we obtain the equivalent maximization problem
\begin{equation}
\max_{\mathcal U}
\; \mathcal J_{\mathrm{BM2PC}}(\mathcal U)  \label{BM2PC prob}
\end{equation}  
where
\begin{equation}
\begin{aligned}
\mathcal J_{\mathrm{BM2PC}}(\mathcal U)
&=
\sum_{k=1}^{n}
\left[
-\lambda^{-1}J_T\!\left(\bm{u}^{(k)}_{0:T-1}\right)
+
\log p\!\left(\bm{u}^{(k)}_{0:T-1}\right)
\right] \\
&
+
\beta
\sum_{i<j}
\left\|
\phi\!\left(\tau(\bm{u}^{(i)}_{0:T-1})\right)
-
\phi\!\left(\tau(\bm{u}^{(j)}_{0:T-1})\right)
\right\|_2^2 . \label{f_BM2PC}
\end{aligned}
\end{equation}
Thus, the first summation in \(\mathcal J_{\mathrm{BM2PC}}\) represents the log posterior density of each candidate sequence, while the second term encourages diversity among the predicted trajectories.

\subsubsection{Sampled Set Optimization}
\label{Sampled Set Optimization}
Directly optimizing \eqref{BM2PC prob} over the continuous control-sequence space is intractable for online MPC. 
At each optimization iteration $\ell$, we define the control prior
as an equally weighted Gaussian mixture centered at the control
modes from the preceding iteration:
\begin{equation}
p_\ell(\bm{u}_{0:T-1})
=
\frac{1}{n}\sum_{k=1}^{n}
\mathcal{N}\left(
\bm{u}_{0:T-1};
\bm{u}_{0:T-1}^{(k,\ell-1)},
\sigma^2 I
\right).
\end{equation}
Here, $\sigma$ is fixed, while the mixture centers are updated
after each optimization iteration. This distribution is used
both to generate control candidates and to evaluate their prior
probabilities. For simplicity, $p(\bm{u}_{0:T-1})$ in the
preceding formulation denotes the prior at the current iteration.
We construct a finite candidate set $\mathcal U_N$ by sampling control sequences from the adaptive Gaussian mixture prior and restrict the search to subsets of this set. This sampling strategy concentrates computation around previously identified control modes, while the prior probability and rollout cost jointly determine the posterior quality of each candidate.

Under this restriction, the continuous set optimization problem is approximated by the finite-cardinality set optimization problem
\begin{equation}
\label{problem in Sampled Set Optimization}
\max_{\mathcal U \subset \mathcal U_N,\; |\mathcal U|=n}
\mathcal J_{\mathrm{BM2PC}}(\mathcal U).
\end{equation}
The objective remains the same as in \eqref{BM2PC prob}, while the feasible domain is replaced by the finite sampled set \(\mathcal U_N\). 
This problem requires selecting \(n\) elements from \(N\) candidates and therefore contains \(\binom{N}{n}\) possible subsets. 
Although exhaustive evaluation is possible for very small \(n\), the number of subsets grows combinatorially with the number of requested modes, motivating the incremental update procedure introduced next.

\subsubsection{Incremental Multimodal Update}
\label{Incremental Multimodal Update}
To avoid combinatorial enumeration, we approximate the finite-cardinality set optimization problem using an incremental greedy procedure. Starting from an empty set, the procedure adds one candidate at each step by maximizing the marginal contribution to the BM2PC objective.

At the \(m\)-th step, given the previously selected set \(\mathcal U_{m-1}\), we define the incremental BM2PC objective as
\begin{equation}
\begin{aligned}
f_{\mathrm{Inc}}(\bm{u}_{0:T-1}\mid \mathcal U_{m-1})
&=
-\lambda^{-1}J_T(\bm{u}_{0:T-1})
+
\log p(\bm{u}_{0:T-1}) \\
&\hspace{-1.5em} +
\beta \displaystyle\sum_{\bm{v}\in \mathcal U_{m-1}}
\left\|
\phi(\tau(\bm{u}_{0:T-1}))-\phi(\tau(\bm{v}))
\right\|_2^2 .
\label{inc_obj}
\end{aligned}
\end{equation}
This objective consists of the posterior score of the candidate control sequence and its accumulated diversity contribution with respect to the already selected set.

Starting from \(\mathcal U_0=\emptyset\), at iteration \(m\) we select
\begin{equation}
\bm{u}^{(m)}_{0:T-1}
=
\arg\max_{\bm{u}\in \mathcal U_N\setminus \mathcal U_{m-1}}
f_{\mathrm{Inc}}(\bm{u}\mid \mathcal U_{m-1}),
\label{BM2PC 1}
\end{equation}
and update
\begin{equation}
\mathcal U_m=
\mathcal U_{m-1}\cup\left\{\bm{u}^{(m)}_{0:T-1}\right\},
\label{BM2PC 2}
\end{equation}
until \(n\) control sequences have been selected.

For any ordered set \(\mathcal U_n=\{\bm{u}^{(1)}_{0:T-1},\dots,\bm{u}^{(n)}_{0:T-1}\}\) generated by this procedure, the BM2PC objective can be decomposed as
\begin{equation}
\mathcal J_{\mathrm{BM2PC}}(\mathcal U_n)
=
\sum_{k=1}^{n}
f_{\mathrm{Inc}}\!\left(\bm{u}^{(k)}_{0:T-1}\mid \mathcal U_{k-1}\right).
\end{equation}
Thus, each incremental step selects a new candidate based on its cost, prior probability, and diversity relative to the selected modes. In practice, min-max normalization is applied to the cost and accumulated diversity scores to balance their scales. For nonnegative diversity weights, the original set objective and its fixed-scale normalized form are supermodular, as the pairwise diversity term yields nondecreasing marginal contributions. However, they are not necessarily monotone, and the step-dependent normalization used in the incremental update does not admit a standard greedy approximation guarantee. The proposed update replaces combinatorial subset enumeration with $n$ sequential candidate selections. Its solution quality and computational efficiency are evaluated against exhaustive set optimization in Sec.~\ref{sec:greedy_vs_exhaustive}.

The rollout costs and prior-density terms in \eqref{inc_obj} are already available for the \(N\) sampled candidates, and the trajectory-feature distances can be precomputed once in parallel. 
Each greedy step then only requires accumulating the corresponding distance scores and evaluating an \(\arg\max\) over the candidate pool. 
Therefore, the multimodal update step introduces only modest additional overhead compared with rollout simulation and collision checking.

\subsubsection{BM2PC algorithm}

\begin{algorithm}[t]
\caption{One inference of the BM2PC Optimizer}
\textbf{Input:} Current state $\bm{x}_0$, previous updated control modes
$\tilde{\mathcal{U}}^{\mathrm{pre}}$, number of candidates $N$,
number of modes $n$, number of optimization iterations $I$\\
\textbf{Output:} Updated control modes $\tilde{\mathcal{U}}$,
executed control $\bm{u}_{\mathrm{exec}}$

\begin{algorithmic}[1]
    \State $\mathcal{U}^{(0)} \gets \tilde{\mathcal{U}}^{\mathrm{pre}}$

    \For{$\ell = 1,\dots,I$} \Comment{Optimization iterations}

        \State
        $\mathcal{U}_N^{(\ell)}
        =
        \left\{\bm{u}^{(i)}_{0:T-1}\right\}_{i=1}^{N}
        \sim
        p(\bm{u}_{0:T-1}\mid\mathcal{U}^{(\ell-1)})$

        \For{each candidate $\bm{u}^{(i)}_{0:T-1}\in\mathcal{U}_N^{(\ell)}$}
            \State Roll out the system from $\bm{x}_0$ under
            $\bm{u}^{(i)}_{0:T-1}$
            \State Collision detection
            \State Compute rollout cost
            $J_T(\bm{u}^{(i)}_{0:T-1})$
        \EndFor

        \State Initialize $\mathcal{U}_0^{(\ell)}\gets\emptyset$

        \For{$m=1,\dots,n$} \Comment{Multimodal update}
            \State
            $\bm{u}^{(m)}_{0:T-1}
            \gets
            \arg\max_{\bm{u}\in
            \mathcal{U}_N^{(\ell)}
            \setminus\mathcal{U}_{m-1}^{(\ell)}}
            f_{\mathrm{Inc}}
            (\bm{u}\mid\mathcal{U}_{m-1}^{(\ell)})$

            \State
            $\mathcal{U}_m^{(\ell)}
            \gets
            \mathcal{U}_{m-1}^{(\ell)}
            \cup
            \{\bm{u}^{(m)}_{0:T-1}\}$
        \EndFor

        \State $\mathcal{U}^{(\ell)}\gets\mathcal{U}_n^{(\ell)}$
    \EndFor

    \State
    $\bm{u}_{\mathrm{exec}}
    \gets
    \text{SelectAction}(\mathcal{U}^{(I)})$

    \State
    $\tilde{\mathcal{U}}
    \gets
    \text{ShiftUpdate}(\mathcal{U}^{(I)})$
\end{algorithmic}
\label{alg:BM2PC}
\end{algorithm}

Algorithm~\ref{alg:BM2PC} summarizes one receding-horizon update of BM2PC. Each MPC update performs $I$ optimization iterations.
At the first iteration, the shifted multimodal control set $\tilde{\mathcal{U}}^{\mathrm{pre}}$ from the previous MPC step is used as the warm start.
At each optimization iteration, we draw $N/n$ control-sequence candidates from a Gaussian centered at each of the $n$ current control modes. The sampled candidates are then rolled out in parallel to obtain their trajectories and costs, and the resulting multimodal set is used as the sampling center for the next optimization iteration.

Before the multimodal update stage, the $K$ lowest-cost candidates are retained to reduce the size of the candidate pool. Within this pool, candidates whose costs exceed a threshold are excluded from the greedy multimodal update. To avoid constructing a full pairwise distance matrix, trajectory-feature distances are computed on demand in parallel between all retained candidates and each newly selected mode, so that each incremental update only requires lightweight distance accumulation and an $\arg\max$ evaluation.

Once the final multimodal control set $\mathcal{U}^{(I)}$ is obtained, one control sequence is selected for execution, and its first control input is applied to the system. In practice, the execution sequence can be chosen deterministically as the minimum-cost sequence (used in our experiments) or sampled probabilistically from a softmax distribution over negative rollout costs. The resulting multimodal set is then shifted to form the warm start for the next MPC step. In our experiments, we adopt a simple shift strategy: each control sequence is shifted forward by one step, and the final vacant entry is filled by copying the second-to-last control input.

    
        

\subsection{DLO Model}
\label{sec:rope_model}

\begin{figure}[!b]
    \centering
    \includegraphics[width=1.0\linewidth]{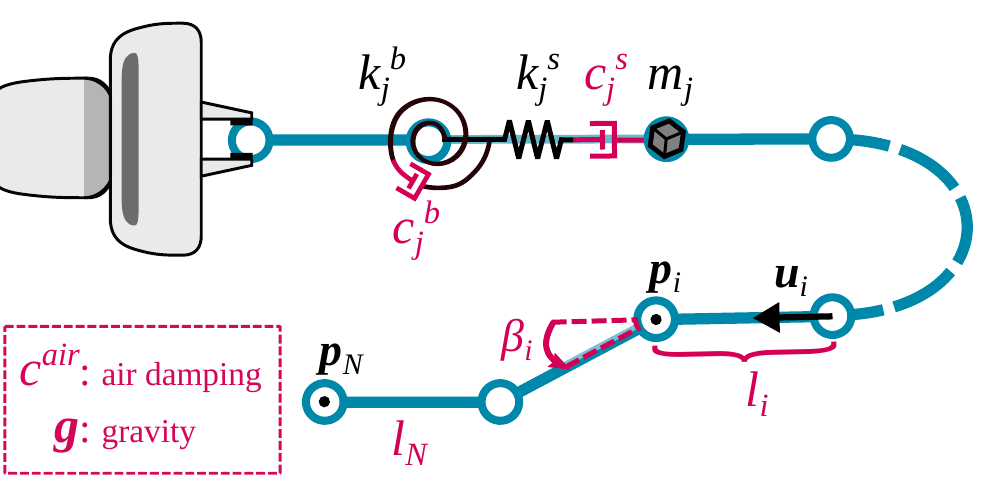}
    \vspace{-0.1cm}
    \caption{\textbf{The DLO model.} The rope is represented by \(N{+}1\) mass points $m_j$ connected by linear spring--damper elements ($k^s_j,c^s_j$) and bending spring--damper elements ($k^b_j,c^b_j$). Gravity $\bm{g}$ and air damping $c^{air}$ are also included.}
    \label{fig:mass-spring-model}
    \vspace{-0.4cm}
\end{figure}

We adopt a lightweight mass–spring–damper model to enable fast rollout evaluation in sampling-based MPC. Following similar works \cite{lv2017physically,long2026self}, the DLO is discretized into \(N+1\) point masses $\{m_i\}_{i=0}^{N}$, connected by \(N\) links, as illustrated in Fig.~\ref{fig:mass-spring-model}. 
Each link is associated with linear stiffness $k_i^s$, linear damping $c_i^s$, bending stiffness $k_i^b$, and bending damping $c_i^b$. For a homogeneous rope, we use spatially uniform material parameters, i.e., $k_i^s=k^s$, $c_i^s=c^s$, $k_i^b=k^b$, and $c_i^b=c^b$ for all valid \(i\).


The full state of a DLO is expressed by \(\bm{X}=[\bm{s}_0^\mathsf{T},\dots,\bm{s}_N^\mathsf{T}]^\mathsf{T}\). 
The \(i\)-th mass point has state \(\bm{s}_i=[\bm{p}_i^\mathsf{T},\bm{v}_i^\mathsf{T}]^\mathsf{T}\), where \(\bm{p}_i,\bm{v}_i\in\mathbb{R}^3\) denote its position and velocity, respectively. 
For the \(i\)-th link, we define \(\bm{l}_i=\bm{p}_i-\bm{p}_{i-1}\), \(l_i=\|\bm{l}_i\|\), and \(\bm{u}_i=\bm{l}_i/l_i\). 
The corresponding rest length is denoted by \(l_i^0\).
The bending angle between adjacent links is \(\beta_i=\arccos(\bm{u}_i^\mathsf{T}\bm{u}_{i+1})\).

For compactness, the following force expressions are written for interior mass points; terms involving indices outside the valid range are omitted at boundary points. 
Please refer to~\cite{long2026self} for more details of the adopted modeling approach.

The linear spring--damper force acting on mass point \(i\) is
\begin{equation*}
\begin{aligned}
\bm{F}_i^{s}
=&-k^s(l_i-l_i^0)\bm{u}_i
+k^s(l_{i+1}-l_{i+1}^0)\bm{u}_{i+1} \\
&-c^s\big((\bm{v}_i-\bm{v}_{i-1})^\mathsf{T}\bm{u}_i\big)\bm{u}_i
+c^s\big((\bm{v}_{i+1}-\bm{v}_i)^\mathsf{T}\bm{u}_{i+1}\big)\bm{u}_{i+1}.
\end{aligned}
\end{equation*}

The bending spring force is modeled as the restoring force induced by bending deformation:
\begin{equation*}
\begin{aligned}
\bm{F}_i^{b}
=&
\frac{k^b\beta_{i-1}}{l_i}
\frac{\bm{u}_{i-1}-\bm{u}_{i}\cos\beta_{i-1}}{\sin\beta_{i-1}}
+
\frac{k^b\beta_i}{l_{i+1}}
\frac{\bm{u}_{i+1}\cos\beta_i-\bm{u}_{i}}{\sin\beta_i} \\
&-
\frac{k^b\beta_i}{l_i}
\frac{\bm{u}_{i}\cos\beta_i-\bm{u}_{i+1}}{\sin\beta_i}
+
\frac{k^b\beta_{i+1}}{l_{i+1}}
\frac{\bm{u}_{i+1}\cos\beta_{i+1}-\bm{u}_{i+2}}{\sin\beta_{i+1}} .
\end{aligned}
\end{equation*}

To account for dynamic deformation, we further include bending damping. The bending angle rate is
\begin{equation*}
\begin{aligned}
\dot{\beta}_i
=&
\left(\frac{\bm{v}_i-\bm{v}_{i-1}}{l_i}\right)^\mathsf{T}
\frac{\bm{u}_i\cos\beta_i-\bm{u}_{i+1}}{\sin\beta_i} \\
&+
\left(\frac{\bm{v}_{i+1}-\bm{v}_i}{l_{i+1}}\right)^\mathsf{T}
\frac{\bm{u}_{i+1}\cos\beta_i-\bm{u}_{i}}{\sin\beta_i},
\end{aligned}
\end{equation*}
and the corresponding damping force is
\begin{equation*}
\begin{aligned}
&\bm{F}_i^{db}
=
\frac{c^b\dot{\beta}_{i-1}}{l_i}
\frac{\bm{u}_{i-1}-\bm{u}_i\cos\beta_{i-1}}{\sin\beta_{i-1}}
+
\frac{c^b\dot{\beta}_i}{l_{i+1}}
\frac{\bm{u}_{i+1}\cos\beta_i-\bm{u}_{i}}{\sin\beta_i} \\
&-
\frac{c^b\dot{\beta}_i}{l_i}
\frac{\bm{u}_i\cos\beta_i-\bm{u}_{i+1}}{\sin\beta_i}
+
\frac{c^b\dot{\beta}_{i+1}}{l_{i+1}}
\frac{\bm{u}_{i+1}\cos\beta_{i+1}-\bm{u}_{i+2}}{\sin\beta_{i+1}} .
\end{aligned}
\end{equation*}


The external force consists of gravity and air damping:
\begin{equation*}
\bm{F}_i^{ext}=m_i\bm{g}-c^{air}\bm{v}_i .
\label{eq:external_force_simple}
\end{equation*}

The acceleration of each free mass point is obtained from Newton's second law:
\begin{equation}
\bm{a}_i=
\frac{
\bm{F}_i^s+
\bm{F}_i^b+
\bm{F}_i^{db}+
\bm{F}_i^{ext}
}{m_i},
\quad i=1,\dots,N .
\label{eq:dlo_acceleration}
\end{equation}
The state is integrated using the symplectic Euler scheme:
\begin{equation}
\begin{aligned}
\bm{v}_i^{t+1} &= \bm{v}_i^t+\Delta t\,\bm{a}_i^t,\\
\bm{p}_i^{t+1} &= \bm{p}_i^t+\Delta t\,\bm{v}_i^{t+1}.
\end{aligned}
\label{eq:symplectic_euler}
\end{equation}

In our manipulation setup, one end of the rope is attached to the robot end-effector (EE). 
The planner outputs a desired Cartesian acceleration command for the EE, denoted by \(\bm{u}_{\mathrm{ctrl}}^t\in\mathbb{R}^3\). 
In the nominal rope model, we assume ideal command execution and use this command as the boundary acceleration of the rope:
\begin{equation}
\bm{a}_0^t=\bm{u}_{\mathrm{ctrl}}^t .
\label{eq:acc_control}
\end{equation}
Accordingly, the endpoint state is updated as
\begin{equation}
\begin{aligned}
\bm{v}_0^{t+1} &= \bm{v}_0^t+\Delta t\,\bm{u}_{\mathrm{ctrl}}^t,\\
\bm{p}_0^{t+1} &= \bm{p}_0^t+\Delta t\,\bm{v}_0^{t+1}.
\end{aligned}
\label{eq:endpoint_update}
\end{equation}
Combining this boundary acceleration update with the dynamics of the free mass points gives the nominal discrete controlled rope model
\begin{equation}
\bm{X}^{t+1}
=
\bm{f}^{\mathrm{rope}}_{\theta,\Delta t}(\bm{X}^{t},\bm{u}_{\mathrm{ctrl}}^{t}),
\label{eq:rope_dynamics}
\end{equation}
where \(\theta=\{m_i,l_i^0,k^s,c^s,k^b,c^b,c^{air}\}\) collects the physical parameters of the rope model.

On real hardware, however, the executed EE motion may deviate from the desired command due to low-level tracking errors and unmodeled execution dynamics. 
To account for this mismatch during planning, we introduce an execution dynamics model in Sec.~\ref{sec:ee_tracking_model}.

\subsection{Execution Dynamics Model}
\label{sec:ee_tracking_model}


\begin{figure}[!b]
    \centering
    \includegraphics[width=1.0\linewidth]{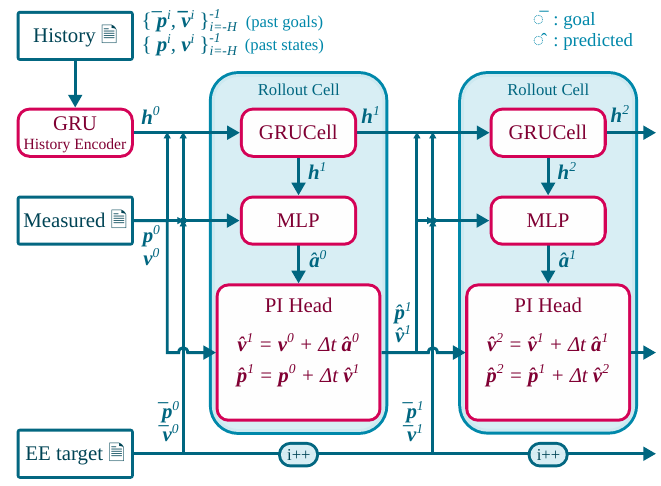}
    \vspace{-0.1cm}
    \caption{\textbf{Execution dynamics model.}
A GRU-based history encoder processes recent measured and desired end-effector states to initialize a latent representation of the current execution dynamics. Conditioned on this latent state and the desired future trajectory, the rollout cell recursively predicts future EE motion. At each rollout step, an MLP predicts the executed acceleration, which is integrated by a physics-informed (PI) head to obtain the next velocity and position.
}
    \label{franka}
    \vspace{-0.4cm}
\end{figure}

Accurate control performance often relies on precise low-level controllers. In practice, however, commanded motions are rarely executed perfectly due to hardware limitations and unmodeled execution dynamics. To cope with this mismatch, we introduce a learned execution dynamics model for predictive compensation of execution errors during planning. 
Our planning method outputs Cartesian acceleration commands, which are integrated into desired EE position trajectories and tracked by a low-level position controller. As illustrated in Fig.~\ref{franka}, the proposed model predicts the actual EE motion from the desired trajectory and recent tracking history.
For notational simplicity, all state variables in this subsection refer to the robot EE.

\textbf{Model structure.} 
A GRU-based history encoder first processes a short history of the measured and desired EE states, $\{\bm{p}^i, \bm{v}^i\}_{i=-H}^{-1}$ and $\{\bar{\bm{p}}^i, \bar{\bm{v}}^i\}_{i=-H}^{-1}$ to initialize a latent representation $\bm{h}_0$ of the current execution dynamics. 
Conditioned on this latent state, the rollout cell is initialized with the current measured EE state $\bm{p}^{0}, \bm{v}^{0}$ and recursively rolls out future EE states conditioned on the desired future trajectory.
The rollout cell consists of a GRUCell, an acceleration decoder (MLP), and a physics-informed integration head (PI Head).
At each rollout step, the GRUCell updates the latent state using the current predicted EE state $\hat{\bm{p}}^{t}$, $\hat{\bm{v}}^{t}$ together with the corresponding desired EE state $\bar{\bm{p}}^{t}$, $\bar{\bm{v}}^{t}$.
The acceleration decoder then predicts the executed acceleration
$\hat{\bm a}^{t}$
from the updated latent state $\bm{h}_{t+1}$ together with the GRUCell input
$(\hat{\bm p}^{t},\hat{\bm v}^{t},
\bar{\bm p}^{t},\bar{\bm v}^{t})$.
Finally, the predicted acceleration is explicitly integrated forward to recover the next EE velocity $\hat{\bm{v}}^{t+1}$ and position $\hat{\bm{p}}^{t+1}$, which are recursively fed into the next rollout step. 
With the integration head, the model predicts only a single variable, i.e., the executed acceleration, instead of directly predicting both $\hat{\bm{p}}^{t+1}$ and $\hat{\bm{v}}^{t+1}$, while preserving their physical relationship.


\textbf{Dataset.}
To collect training data, the robot was commanded to execute randomized motions covering diverse acceleration profiles and motion patterns, resulting in approximately 1.2 million synchronized state samples recorded \highlight{at 1\,kHz}. The randomized motions were generated using piecewise-constant Cartesian acceleration commands, with each component independently sampled from a uniform distribution. To excite a broad range of execution dynamics, the robot cyclically alternated between slow, medium, and fast motion segments, \highlight{with acceleration magnitudes up to 1.5, 3, and 5\,m/s$^2$ and segment durations of 300--800\,ms, 120--300\,ms, and 30--120\,ms, respectively.} Position and velocity limits were enforced throughout data collection to keep the robot within a predefined workspace while ensuring diverse motion patterns.

\textbf{Training.}
The recorded trajectories were then decomposed into overlapping short rollout windows, each consisting of a history segment and a future trajectory segment. The model was trained autoregressively using multi-step rollout prediction, where the predicted EE positions and velocities were supervised by the measured trajectories over the entire rollout horizon using a weighted mean-squared-error loss. To avoid data leakage from overlapping windows, training and test sets are constructed from independently recorded trajectories, with windows generated separately for each set.

\subsection{Collision Detection}
Obstacles are modeled as cylinders with known length and radius, allowing collision checks between the discretized rope nodes and obstacles to be computed analytically.
This procedure corresponds to the collision detection module in Fig.~\ref{fig:method-overview}.
For computational efficiency, collision checking is performed on the rope nodes rather than continuously along each rope segment.
To reduce the possibility of missed collisions between adjacent nodes, the obstacle geometry used for planning is conservatively inflated by a safety margin.
If any rope node intersects the inflated obstacle, the corresponding rollout is assigned an infinite cost and excluded from the feasible candidate set; otherwise, the rollout is treated as collision-free for planning.
Combined with the dense discretization of the rope, this lightweight approximation provides reliable obstacle avoidance in our experiments while introducing negligible computational overhead.

\section{Experiments and Results}
We evaluate BM2PC on dynamic rope manipulation in both simulation and hardware. Trajectory tracking and disturbance rejection assess its dynamic control capability using a single mode ($n=1$), while obstacle avoidance evaluates its multimodal planning capability ($n=3$) when multiple feasible control strategies exist. We further examine the contributions of multimodal maintenance and diversity promotion, as well as the accuracy and computational efficiency of the incremental update.

\subsection{Experimental setup}
\subsubsection{Cost function} We use a simple $\ell_1$ distance between the rollout rope-tip trajectory and the corresponding target trajectory as the loss function. 
Rollouts that result in collisions between the rope and obstacles are assigned an infinite loss.

\subsubsection{Baseline Controllers}

We compare the proposed controller against three baseline methods: SPiD~\cite{long2026self}, \mbox{C-MPPI}~\cite{patrick2024path}, and SVMPC~\cite{lambert2020stein}. 

SPiD is a physics-informed self-supervised learning framework for deformable object manipulation, where a neural controller is jointly trained with a differentiable deformable object model through task-oriented optimization.

C-MPPI and SVMPC are both sampling-based multimodal MPC methods. \mbox{C-MPPI} is a clustering-based extension of Model Predictive Path Integral (MPPI) control, which uses DBSCAN to cluster rollout trajectories into different modes before control update computation. In contrast, SVMPC formulates model predictive control as a Bayesian inference problem and employs Stein Variational Gradient Descent (SVGD) to optimize a multimodal particle-based posterior distribution over control trajectories.

We implemented all baselines following the descriptions in the corresponding papers.

\subsubsection{Validation scenarios}
To comprehensively evaluate the effectiveness, multimodal capability, and robustness of BM2PC, we design three groups of experiments under different settings.

(1) \textbf{Trajectory Tracking:} This is the primary evaluation task, where the controller is required to track target trajectories with different shapes and execution durations.

(2) \textbf{Obstacle Avoidance:} Moving obstacles are placed along the target trajectory to evaluate the controller's multimodal planning capability under dynamic environments.

(3) \textbf{Disturbance Rejection:} Random external force perturbations are applied to the rope during motion to evaluate the robustness of the controller against unexpected disturbances.

\subsection{Simulation Experiments}
\textbf{Environment setup.} The simulation experiments are conducted in MuJoCo 3.2.2 using a 1\,m rope model composed of 40 serially connected rigid segments. 
The proposed controller outputs target accelerations for the rope's top endpoint at 40\,Hz in three-dimensional space. These accelerations are integrated into target positions and executed by a high-gain low-level PD controller running at 1\,kHz, resulting in negligible command-execution mismatch and therefore eliminating the need for the execution dynamics model in simulation.
The full rope state is obtained through the MuJoCo Python interface by accessing the position and velocity of all rope segment bases.

The controller operates at 40\,Hz, while the MuJoCo simulation is
integrated at 1\,kHz. The BM2PC and controller parameters used in the three simulation experiments are summarized in Table~\ref{tab:bm2pc_hyperparameters}.

The computational bottleneck of BM2PC is rollout simulation, which must remain highly efficient to enable large rollout batches and long prediction horizons under real-time control constraints. 
We explored several mainstream GPU computation frameworks, including PyTorch, JAX, Taichi~\cite{hu2019difftaichi}, and NVIDIA Warp~\cite{macklin2022warp}, and found Warp to provide significantly faster rollout simulation than the other alternatives. 
Accordingly, the simulation model was implemented using NVIDIA Warp. 
Since the library is still under active development and lacks mature support for several neural network operators, modules used in the execution dynamics model, including MLPs and GRUs, were manually implemented.

\subsubsection{Trajectory Tracking}\label{sec:exp-sim-traj}

\begin{table}[t]
\centering
\caption{BM2PC and controller parameters used in the simulation experiments.}
\label{tab:bm2pc_hyperparameters}
\small
\setlength{\tabcolsep}{3pt}
\renewcommand{\arraystretch}{1.08}

\begin{tabular}{lccc}
\toprule
Parameter & Tracking & Obstacle Avoid. & Dist. Rej. \\
\midrule
Prediction horizon $H$
& 30 & 35 & 35 \\

Number of modes $n$
& 1 & 3 & 1 \\

Samples / iteration $N$
& 402 & 402 & 402 \\

Optimization iterations $I$
& 10 & 1 & 10 \\

Sampling std. $\sigma$
& 1.5 & 1.5 & 1.5 \\

Sampling covariance $\Sigma$
& $\sigma^2 I$ & $\sigma^2 I$ & $\sigma^2 I$ \\

Cost weight $\lambda^{-1}$
& $1.7{\times}10^5$ & 850 & $1.7{\times}10^5$ \\

Diversity weight $\beta$
& 600 & 600 & 600 \\

Retained candidates $K$
& 200 & 200 & 200 \\

Filtering threshold
& $10^5$ & $10^5$ & $10^5$ \\

Acceleration limits (m/s$^2$)
& $\pm5$ & $\pm5$ & $\pm5$ \\

MPC frequency (Hz)
& 40 & 40 & 40 \\

Low-level frequency (Hz)
& 1000 & 1000 & 1000 \\
\bottomrule
\end{tabular}
\end{table}

\begin{figure*}[h]
    \centering
    \includegraphics[width=\textwidth]{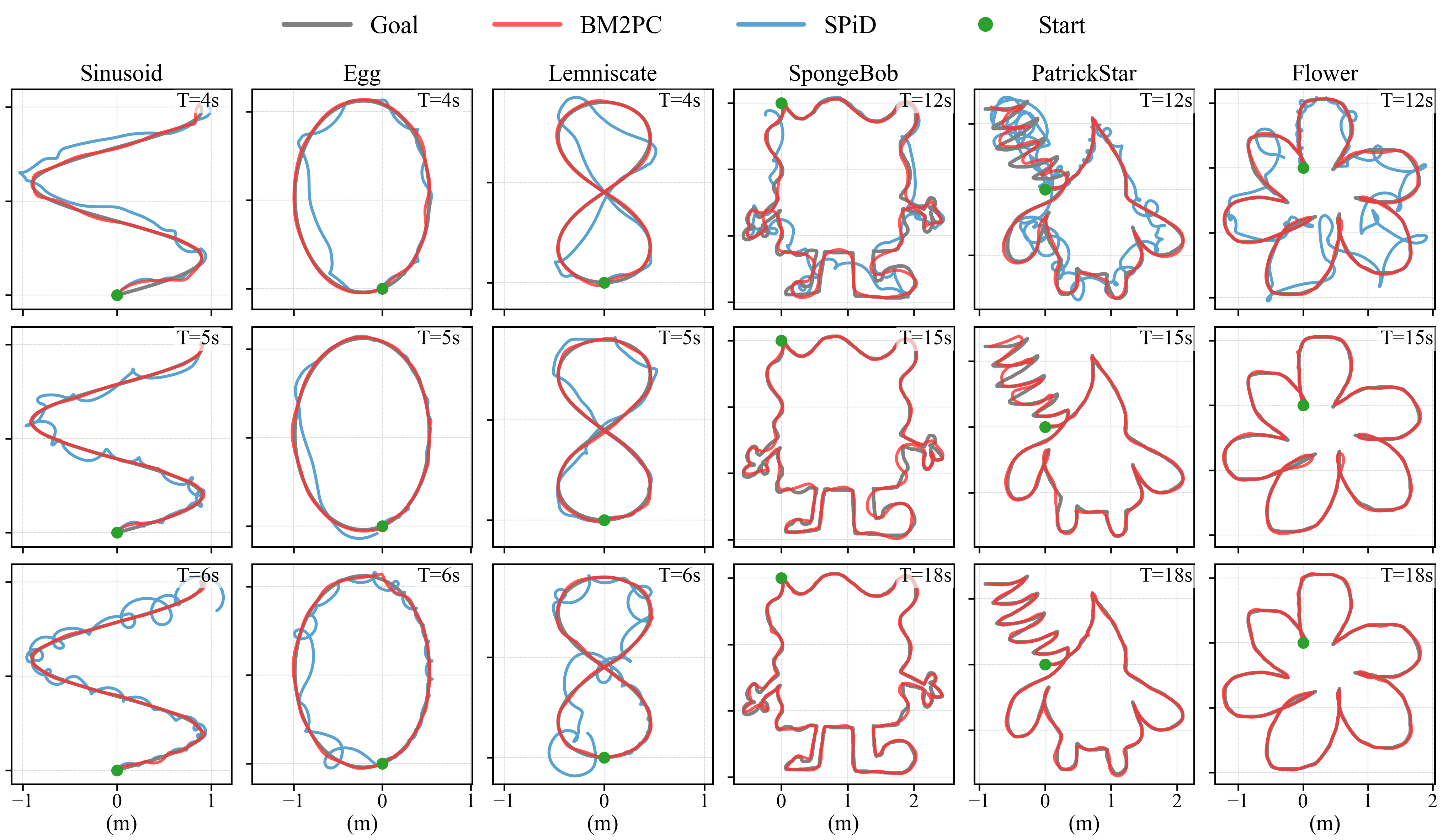}
    \vspace{-0.6cm}
    \caption{\textbf{Trajectory tracking results (Sim).} Comparison of trajectory tracking performance of the proposed BM2PC method and the SPiD baseline on both smooth parametric trajectories (Sinusoid, Egg, and Lemniscate) and complex hand-drawn trajectories (SpongeBob, PatrickStar, and Flower) under different execution durations. The proposed method achieves accurate tracking across most trajectories, while SPiD fails to reproduce fine geometric details, particularly on complex trajectories with sharp turns and rapid local variations. 
    Thus, SPiD is not shown in the longer cases ($T{>}12$s) to avoid clutter.
}
    \label{fig:sim tracking}
\end{figure*}

\begin{table*}[t]
\centering
\caption{Trajectory tracking simulation task specifications and tracking RMSE comparison.}
\label{tab:sim-tracking}

\footnotesize
\setlength{\tabcolsep}{3.1pt}
\renewcommand{\arraystretch}{1.15}

\begin{tabular}{@{}lcccccccccccccccccc@{}}
\toprule

\multirow{2}{*}{\shortstack{Trajectory\\(Length)}}
& \multicolumn{3}{c}{\textbf{Sinusoid}}
& \multicolumn{3}{c}{\textbf{Egg}}
& \multicolumn{3}{c}{\textbf{Lemniscate}}
& \multicolumn{3}{c}{\textbf{SpongeBob}}
& \multicolumn{3}{c}{\textbf{PatrickStar}}
& \multicolumn{3}{c}{\textbf{Flower}} \\

& \multicolumn{3}{c}{\textbf{(5.085 m)}}
& \multicolumn{3}{c}{\textbf{(5.844 m)}}
& \multicolumn{3}{c}{\textbf{(5.540 m)}}
& \multicolumn{3}{c}{\textbf{(18.3 m)}}
& \multicolumn{3}{c}{\textbf{(18.1 m)}}
& \multicolumn{3}{c}{\textbf{(17.5 m)}} \\

\cmidrule(lr){2-4}
\cmidrule(lr){5-7}
\cmidrule(lr){8-10}
\cmidrule(lr){11-13}
\cmidrule(lr){14-16}
\cmidrule(lr){17-19}

Duration (s)
& 4 s & 5 s & 6 s
& 4 s & 5 s & 6 s
& 4 s & 5 s & 6 s
& 12 s & 15 s & 18 s
& 12 s & 15 s & 18 s
& 12 s & 15 s & 18 s \\

Max Velocity (m/s)
& 2.245 & 1.797 & 1.498
& 2.552 & 2.043 & 1.703
& 2.089 & 1.672 & 1.394
& 1.876 & 1.501 & 1.251
& 1.857 & 1.486 & 1.238
& 1.765 & 1.412 & 1.177 \\

\midrule

SPiD RMSE (cm)
& 19.26 & 9.950 & 10.97
& 23.32 & 11.23 & 9.407
& 16.17 & 6.391 & 11.66
& 16.12 & 19.06 & 30.74
& 27.94 & 37.65 & 36.14
& 29.47 & 39.80 & 30.85 \\

BM2PC RMSE (cm)
& \textbf{9.690} & \textbf{6.230} & \textbf{4.650}
& \textbf{6.900} & \textbf{4.850} & \textbf{3.960}
& \textbf{6.570} & \textbf{4.800} & \textbf{3.770}
& \textbf{7.500} & \textbf{5.670} & \textbf{4.080}
& \textbf{8.500} & \textbf{7.000} & \textbf{3.580}
& \textbf{5.430} & \textbf{4.120} & \textbf{3.380} \\

\bottomrule
\end{tabular}

\end{table*}

In this experiment, we evaluate our method on six trajectories. First, we prepared three analytically generated smooth trajectories—\textit{Sinusoid}, \textit{Egg}, and \textit{Lemniscate}—each evaluated under three different execution durations. Since these trajectories are generated from smooth functions, they do not fully capture the challenges of \highlight{diverse} trajectory tracking. To further evaluate the generalization capability of our method, we additionally designed three hand-drawn trajectories: \textit{SpongeBob}, \textit{PatrickStar}, and \textit{Flower}. These trajectories exhibit significantly more complex and irregular characteristics, including sharp corners, abrupt turns, and rapid local geometric variations. Each trajectory was also tested under three different time settings. 
The detailed trajectory specifications are summarized in Table~\ref{tab:sim-tracking}. 
Importantly, all 18 trajectory-tracking tasks use the same Tracking configuration in Table~\ref{tab:bm2pc_hyperparameters}, without trajectory-specific retuning. 
Following the implementation details of the original work~\cite{long2026self}, the SPiD controller outputs velocity commands at 100~Hz, which are subsequently integrated into target positions and executed by a 1~kHz PD controller.

\begin{figure*}[t]
      \centering
      \includegraphics[width=0.95\linewidth]{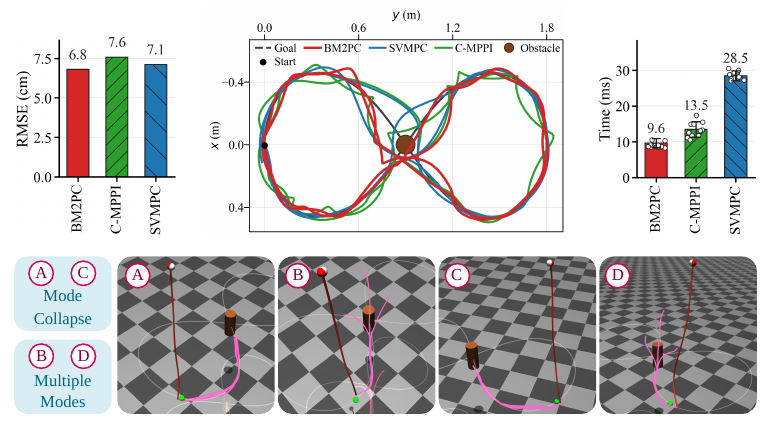}
      \caption{\textbf{Obstacle Avoidance (Sim).} Comparison of BM2PC, \mbox{C-MPPI}, and SVMPC on the obstacle avoidance task. The top row presents the rope-tip trajectories together with the tracking RMSE and average inference time. 
      The bottom row shows key snapshots of BM2PC, illustrating mode collapse in unimodal regions (A,C) and multimodal emergence near the obstacle (B,D).
      The rope tip may pass above the obstacle without touching it, as seen in snapshots B and D.}
      \label{sim obs}
      \vspace{-0.2cm}
\end{figure*}

The trajectory tracking results are shown in Fig.~\ref{fig:sim tracking}. It can be observed that our method achieves accurate tracking performance on nearly all trajectories. The main sources of tracking error are the fine details of the SpongeBob trajectory around the finger regions, which are difficult to reproduce precisely; and the final zigzag pattern in the fastest PatrickStar setting ($T=12$~s), which cannot be fully tracked due to the inherent delay characteristics of the rope dynamics. 
In contrast, the SPiD baseline exhibits substantially inferior performance, despite being evaluated only on the mass-spring-damper DLO model (Sec.~\ref{sec:rope_model}). For the first three smooth trajectories, SPiD achieves reasonable coarse-level tracking performance; however, for the more complex hand-drawn trajectories, it is only capable of following the overall motion trend while failing to reproduce detailed geometric features. Moreover, increasing the execution duration does not consistently reduce the tracking error, since the rope exhibits stronger free oscillations during slower motions. \highlight{For this reason, only the $T=12$~s SPiD results are shown for the three complex trajectories to avoid visual clutter.} As also confirmed by our results, arbitrary trajectory tracking remains particularly challenging for learning-based approaches due to their limited generalization capability and the requirement for large and diverse training datasets.

\subsubsection{Obstacle Avoidance}
\label{sec:Obstacle Avoidance}

The trajectory tracking task is effectively unimodal, with optimal solutions typically concentrated around a single mode. To demonstrate the multimodal capability of BM2PC, we place a cylindrical obstacle with a diameter of 10\,cm and a height of 24\,cm at the center of the Lemniscate trajectory, creating multiple feasible avoidance behaviors. To provide a safety margin during obstacle avoidance, the obstacle is inflated by $2$\,cm in all directions for collision evaluation, resulting in an effective planning diameter of $14$\,cm and an effective height of $28$\,cm.

In this experiment, the multimodal MPC methods \mbox{C-MPPI} and SVMPC are used as baselines. 
Since obstacle avoidance is inherently multimodal, BM2PC maintains $n=3$ control modes, and the prediction horizon is increased from $H=30$ to $H=35$ to provide sufficient look-ahead for collision avoidance.
Similar to BM2PC, both \mbox{C-MPPI} and SVMPC update the target acceleration at 40\,Hz and execute the integrated position commands through the same low-level PD controller running at 1\,kHz.
For a controlled comparison, \mbox{C-MPPI} and SVMPC use the same prediction horizon ($H=35$), sampling budget ($N=402$ per optimization iteration), sampling covariance, and Cartesian acceleration limits ($\pm5\,\mathrm{m/s^2}$).
Both BM2PC and \mbox{C-MPPI} use one optimization iteration per control step, whereas SVMPC uses two SVGD update iterations to achieve satisfactory performance.
The number of modes for BM2PC and SVMPC is set to three, while \mbox{C-MPPI} automatically determines the number of modes from the rollout clustering results.

The obstacle avoidance results are shown in Fig.~\ref{sim obs}. None of the controllers collide with the obstacle during execution. Although the BM2PC trajectory appears to overlap with the obstacle in the 2D projection, the rope tip actually passes over the obstacle in 3D space without collision. 
The bar chart on the left side of Fig.~\ref{sim obs} presents the RMSE over the entire trajectory. The three controllers achieve comparable tracking accuracy, indicating that obstacle avoidance is achieved without significantly deviating from the target trajectory. 
However, a clear difference appears in the inference time. The bar chart on the right side of Fig.~\ref{sim obs} shows the average computation time per inference step. 

BM2PC achieves the lowest inference time, mainly due to its computationally lightweight multimodal update (Alg.~\ref{alg:BM2PC}, lines 10--12), which requires only 1.55\,ms on average.
In contrast, the additional computation time of \mbox{C-MPPI} mainly comes from its multimodal optimization stage, which requires approximately 4.99\,ms on average and increases linearly with the rollout batch size. By comparison, the computation time of the multimodal optimization stage in BM2PC remains nearly unaffected by the rollout batch size.
SVMPC exhibits the highest computation time because its multimodal optimization stage relies on iterative SVGD-based gradient updates, where each update only incrementally refines the particle distribution. As a result, two optimization iterations are required in each inference step to achieve satisfactory performance. We also tested SVMPC with only one update iteration, which approximately halves the computation time, but increases the RMSE to 65.4\,cm.

Key snapshots of BM2PC are shown at the bottom of Fig.~\ref{sim obs}. In subfigures A and C, the rope remains far from the obstacle, reducing the task to standard trajectory tracking. As a result, all modes collapse into a single solution. In contrast, in subfigures B and D, the rope approaches the obstacle, and different modes emerge to maximize behavioral diversity while simultaneously avoiding collisions and maintaining low trajectory costs, resulting in distinct leftward, rightward, and upward avoidance behaviors.

\subsubsection{Disturbance Rejection}
\begin{figure}[t]
      \centering
      \includegraphics[width=0.83\linewidth]{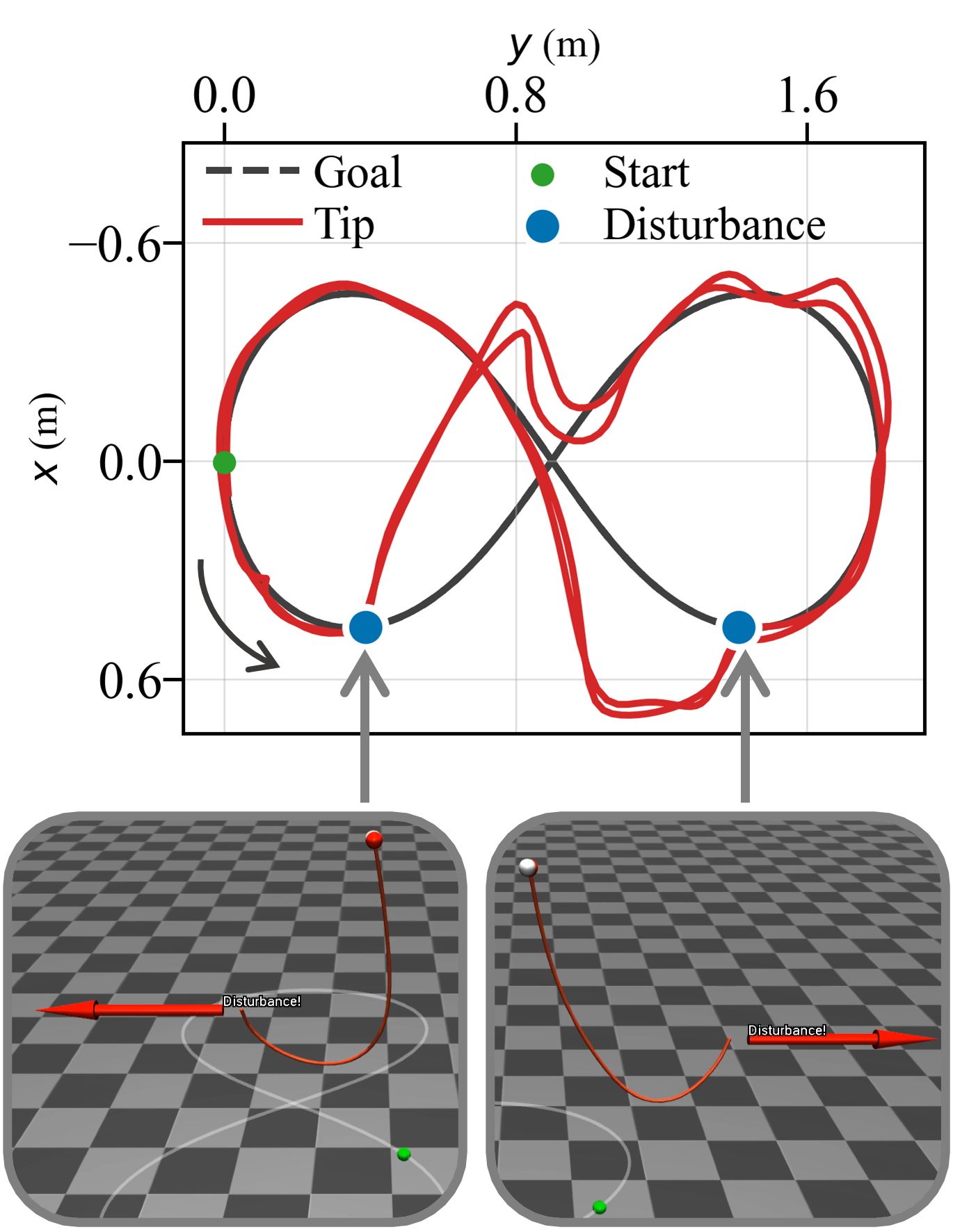}
      \caption{\textbf{Disturbance Rejection (Sim).} The rope tip is subjected to external disturbances at the blue points to evaluate the robustness of the controller. The corresponding rope-tip trajectories and key snapshots of the experiment are presented. BM2PC rapidly suppresses the induced oscillations and drives the rope tip back to the desired trajectory.}
      \label{sim robustness}
      \vspace{-0.4cm}
\end{figure}
To evaluate the robustness of BM2PC against unknown external disturbances, four $1$\,N disturbances are applied to the rope tip at $t=1.5$, $4.0$, $6.5$, and $9.0$\,s, with alternating directions along the $x$ axis and a duration of $50$\,ms each. The rope mass is $100$\,g. As shown in Fig.~\ref{sim robustness}, these disturbances are applied at different phases of the motion and induce substantial transient deformation of the rope, resulting in large deviations of the rope tip from the nominal trajectory.

Across the four disturbance events, the maximum rope-tip deviation ranges from $62.0$ to $73.0$\,cm, with an average of $66.7$\,cm. Despite these substantial deviations, the tracking error returns below $5$\,cm within $0.68$--$0.93$\,s after disturbance onset, with an average recovery time of $0.78$\,s. The post-recovery RMSE remains low, averaging $4.69$\,cm across the four disturbance events. These results indicate that BM2PC can rapidly suppress the induced oscillations and recover accurate trajectory tracking even after large transient deformations.
For this experiment, the prediction horizon is increased from $H=30$ to $H=35$ to provide a longer look-ahead under external disturbances, while all other BM2PC hyperparameters remain the same as in the trajectory-tracking experiments.

\subsection{Ablation and Optimization Analysis}

\subsubsection{Ablation of the Multimodal Update}
\label{sec:ablation}

To evaluate the contributions of multimodal maintenance and diversity promotion, we compare BM2PC ($n=3$, $\beta=600$) with two ablated variants: $n=1$, which maintains a single control mode, and $\beta=0$, which retains three modes without the diversity reward. The latter selects the three highest-scoring candidates based on the cost and prior terms. All variants use the same sampling budget ($N=402$) and planning parameters.

\begin{figure*}[t]
    \centering
    \includegraphics[width=\textwidth]{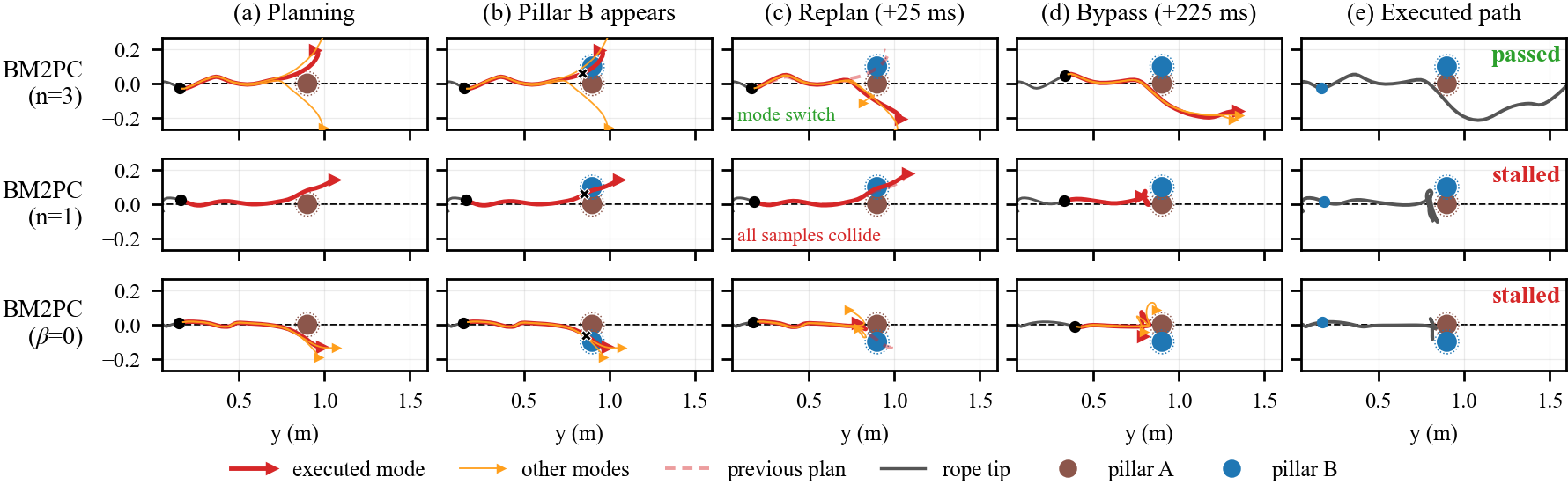}
    \caption{\textbf{Multimodal adaptation under sudden blocking.}
    Comparison of representative trajectories from experiments with $\tau=0.70$\,s.
    (a) Predicted control modes before blocking.
    (b) The second obstacle appears and blocks the executed plan.
    (c) Replanning after 25\,ms.
    (d) Predicted modes after 225\,ms.
    (e) Executed rope-tip trajectories.
    BM2PC switches to an alternative mode on the unblocked side,
    whereas the ablated variants fail to recover a feasible passage
    in this example.}
    \label{fig:ablation_modes}
    \vspace{-0.2cm}
\end{figure*}

To investigate the benefit of multimodal planning under changing constraints, we introduce a \textit{sudden obstacle scenario}. A vertically extended cylindrical obstacle (1\,m high) prevents the rope from passing over it, leaving two alternative routes on either side. A second obstacle suddenly appears on the side preferred by the controller at a lead time $\tau\in[0.65,0.90]$\,s before the reference crossing, blocking its previously selected route. This abrupt change can render the current plan infeasible and cause collisions if the controller cannot find and execute an alternative trajectory in time, particularly due to the delayed dynamic response of the rope. Each variant is evaluated across 10 seeds under identical planning conditions. A trial is considered successful if the rope tip passes the fixed obstacle by at least 10\,cm within 1\,s after the reference crossing, without rope--obstacle contact.

\begin{table}[b]
\centering
\caption{\textbf{Ablation of the multimodal update.}
Results are aggregated over $\tau\in[0.65,0.90]$\,s.}
\label{tab:ablation}
\small
\setlength{\tabcolsep}{4pt}
\renewcommand{\arraystretch}{1.08}
\begin{tabular}{lccc}
\toprule
Metric & BM2PC & $n=1$ & $\beta=0$ \\
\midrule
Success rate (\%)
    & \textbf{64.3} & 45.2 & 47.0 \\
Collision rate (\%)
    & \textbf{25.2} & 39.3 & 35.7 \\
Mean max. tracking error (cm)
    & \textbf{56.2} & 80.2 & 76.6 \\
\bottomrule
\end{tabular}
\end{table}

\begin{figure}[t]
    \centering
    \includegraphics[width=\linewidth]{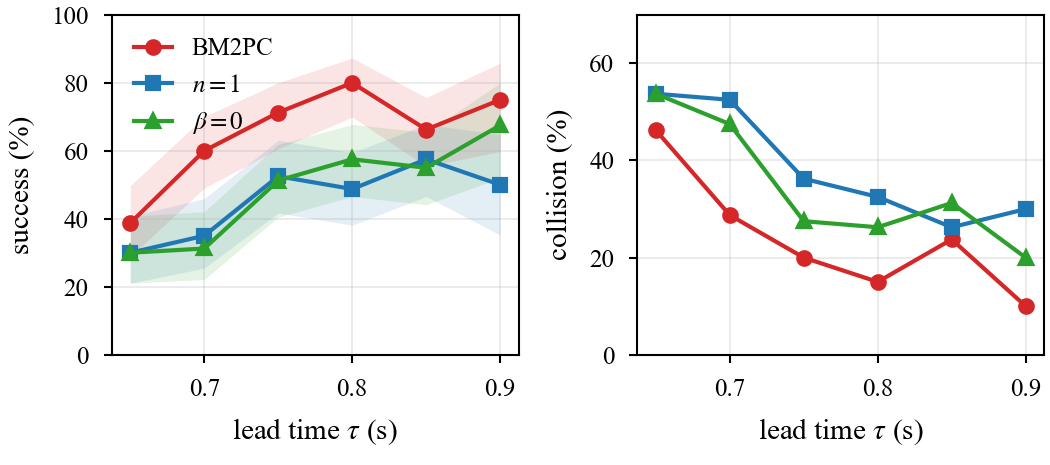}
    \caption{\textbf{Sudden obstacle ablation.}
    Success and collision rates versus obstacle lead time $\tau$.
    Shaded regions indicate 95\% Wilson confidence intervals
    for success rates.}
    \label{fig:ablation_tau}
    \vspace{-0.6cm}
\end{figure}

As shown in Table~\ref{tab:ablation}, BM2PC achieves a 64.3\% success rate, compared with 45.2\% for $n=1$ and 47.0\% for $\beta=0$, while reducing the collision rate from 39.3\% and 35.7\%, respectively, to 25.2\%. 
Fig.~\ref{fig:ablation_tau} further shows that the improvement is particularly pronounced for $\tau=0.70$--$0.80$\,s, where BM2PC exceeds both ablated variants by about 20--30 percentage points in success rate.
Fig.~\ref{fig:ablation_modes} plots some of the trajectories to illustrate the underlying mechanism: BM2PC maintains an alternative mode on the unblocked side and switches to it when the preferred route becomes infeasible. In contrast, $n=1$ lacks an alternative mode, while $\beta=0$ tends to retain similar solutions, making rapid recovery more difficult. These results demonstrate the importance of both multimodal maintenance and diversity promotion for adapting to sudden environmental changes.

\subsubsection{Greedy vs. Exhaustive Set Optimization}
\label{sec:greedy_vs_exhaustive}

To evaluate the efficiency and solution quality of the incremental multimodal update proposed in Sec.~\ref{Incremental Multimodal Update}, we compare BM2PC with exhaustive set optimization formulated in Sec.~\ref{Sampled Set Optimization}, without the greedy approximation. The comparison is conducted on a Lemniscate tracking task with cylindrical obstacles, using the same hyperparameters as the preceding obstacle-avoidance experiment (Sec.~\ref{sec:Obstacle Avoidance}). The exhaustive baseline evaluates all possible three-mode subsets using a fixed pool-normalized version of the Bayesian set objective. The rollout costs are normalized using the minimum and maximum values in the candidate pool, while the pairwise diversity terms are scaled by the maximum pairwise distance. The same reference objective is used to evaluate the sets selected by BM2PC, whose incremental update normalizes accumulated diversity at each selection step.

We evaluate selection quality on 4,375 candidate pools collected from BM2PC across ten random seeds. For each candidate pool, the normalized optimality gap is defined as
\begin{equation}
\eta = \frac{\mathcal J_{\mathrm{BM2PC}}^{\star}-\mathcal J_{\mathrm{BM2PC}}(\mathcal U_n)}{\mathcal J_{\mathrm{BM2PC}}^{\star}-\mathcal J_{\mathrm{BM2PC}}^{\min}},
\label{eq:selection_gap}
\end{equation}
where $\mathcal U_n$ is the set selected by the incremental update in \eqref{BM2PC 2}, and $\mathcal J_{\mathrm{BM2PC}}^{\star}$ and $\mathcal J_{\mathrm{BM2PC}}^{\min}$ denote the maximum and minimum values of the pool-normalized objective derived from \eqref{f_BM2PC} over all three-mode subsets of the same candidate pool, obtained through exhaustive enumeration.

Despite relying on an incremental greedy approximation, BM2PC achieves
a near optimal behaviour
with a mean normalized optimality gap of only 1.14\% and a 95th-percentile gap of 6.80\% in the worse case scenarios. In 56.07\% of the cases, it achieved the exhaustive optimum. In closed-loop evaluations, BM2PC maintains comparable rope-tip tracking accuracy (6.22\,cm versus 6.08\,cm RMSE) while reducing the average planning time from 90.14\,ms to 9.63\,ms, as shown in Table~\ref{tab:incremental_validation}. These results demonstrate that the incremental multimodal update substantially improves computational efficiency while preserving high-quality mode selection and comparable closed-loop control performance.

\begin{table}[t]
\centering
\caption{Incremental multimodal update versus
pool-normalized exhaustive selection.}
\label{tab:incremental_validation}
\small
\setlength{\tabcolsep}{4pt}
\renewcommand{\arraystretch}{1.1}

\begin{tabular}{lcc}
\toprule
Metric & BM2PC & Exhaustive \\
\midrule
Rope-tip RMSE (cm)
    & 6.22$\pm$0.48 & 6.08$\pm$0.53 \\
Planning time (ms)
    & 9.63 & 90.14 \\
Multimodal update (ms)
    & 1.55 & 81.29 \\
\bottomrule
\end{tabular}
\end{table}

\subsection{Real-world experiments}
\textbf{Hardware setup.}
In the real-world experiments, we use a Franka Emika Panda robotic arm equipped with a Franka Hand gripper, which grasps the top end of the rope. The desired end-effector (EE) acceleration is updated by BM2PC at $66.67\,\mathrm{Hz}$ on a $95\,\mathrm{W}$ NVIDIA GeForce RTX 5090 Laptop GPU. Similar to the simulation setup, the acceleration commands are integrated into target positions and executed by a low-level position controller running at $1\,\mathrm{kHz}$. 

In our preliminary trials without the execution dynamics model (Sec.~\ref{sec:ee_tracking_model}), we observed that the low-level controller cannot perfectly track the commanded motion during fast dynamic manipulation, resulting in tracking errors of up to $5\,\mathrm{cm}$. 
This motivated the final design with the execution dynamics model (Fig.~\ref{fig:method-overview}). 

Rope segment positions are tracked at $100\,\mathrm{Hz}$ using an OptiTrack motion capture system consisting of 12 infrared cameras. For this purpose, we attach $9$ equally spaced reflective markers along the rope. The velocity of each marker is estimated from consecutive OptiTrack position measurements using an Alpha-Beta filter, which performs prediction and residual-based correction to reconstruct the full rope state.

The rope used in the experiments is a cotton rope with a linear density of $16\,\mathrm{g/m}$ and a length of $80\,\mathrm{cm}$. To enrich the dynamic characteristics of the system, an additional weight of $15\,\mathrm{g}$ is attached to the rope tip. The obstacle is a 3D-printed cylindrical object with a diameter of $5.5\,\mathrm{cm}$ and a height of $14\,\mathrm{cm}$. To provide a safety margin for collision avoidance, we conservatively inflate the obstacle about its center by $2.25\,\mathrm{cm}$ in the radial direction and $5\,\mathrm{cm}$ in both vertical directions, resulting in an effective planning diameter of $10\,\mathrm{cm}$ and an effective height of $24\,\mathrm{cm}$.

\subsubsection{Execution Dynamics Model Evaluation}

\begin{figure}[t]
      \centering
      \includegraphics[width=1.0\linewidth]{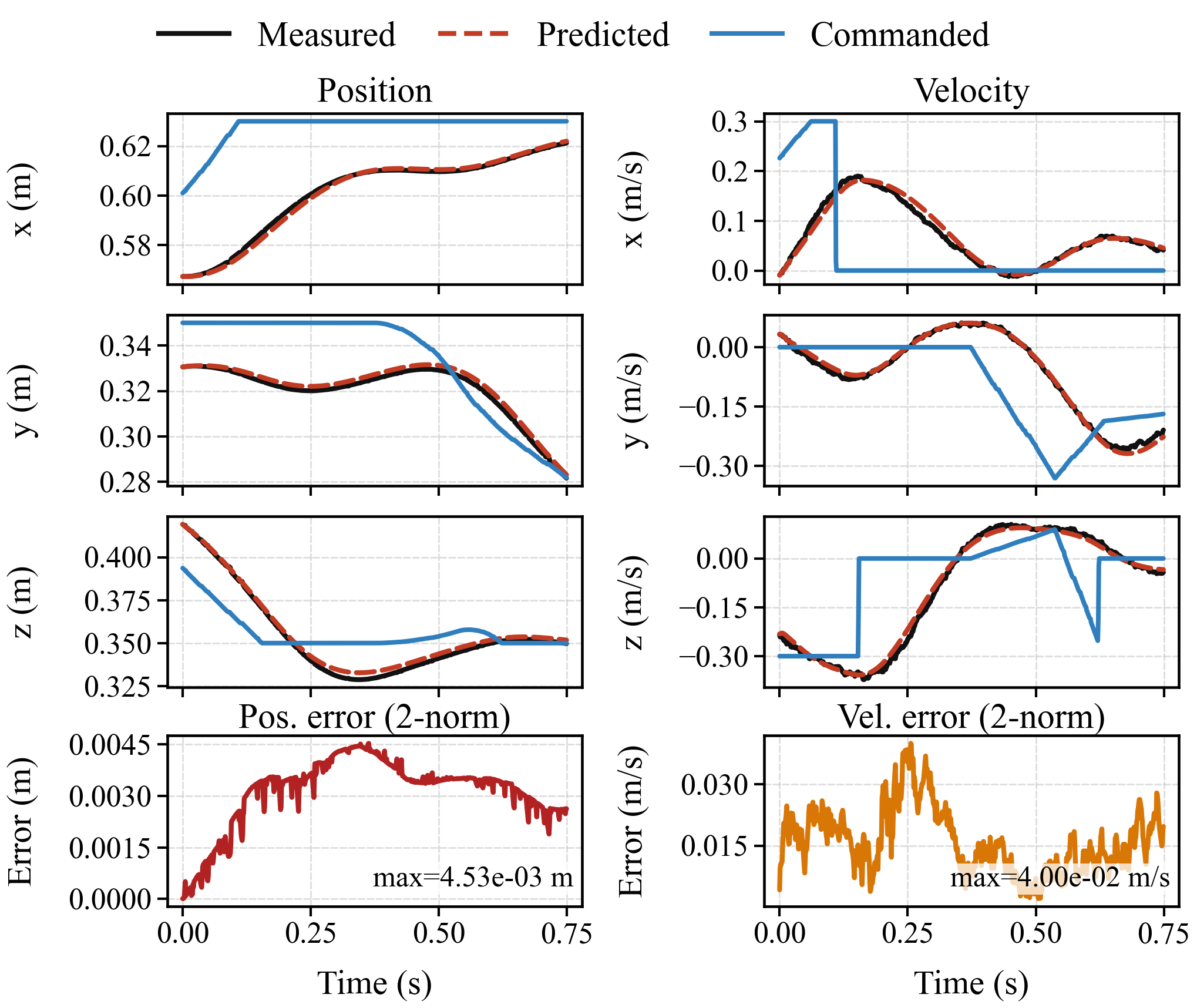}
      \caption{A test-time example of the execution dynamics model.}
      \label{Execution Dynamics Model}
      \vspace{-0.5cm}
\end{figure}

For real-world dynamic manipulation, accurate modeling of the execution dynamics of the Franka robot is essential. 
In our system, the execution dynamics model predicts the future motion of the Franka end-effector over a 0.75\,s horizon. 
The execution dynamics model uses a history of 31 states sampled at 10\,ms intervals, covering 0.3\,s, with each state containing 12 dimensions of measured and commanded EE positions and velocities. The history encoder consists of a single-layer GRU with 64 hidden units, followed by a linear layer (64--64) and tanh activation. The rollout model uses a GRUCell (12-dimensional input, 64 hidden units) and an MLP (76--64--64--3, ReLU activations) to predict acceleration, followed by semi-implicit Euler integration at 1\,ms intervals over 750 steps. The model contains 43.4k parameters. Training uses Adam with a learning rate of $2\times10^{-4}$, batch size 256, gradient clipping at a global norm of 1.0.

The dataset contains approximately 1.5 million state samples, with 0.25 million reserved for testing and the remaining 1.25 million split into training (85\%) and validation (15\%) sets.
On real trajectory test data, the learned execution dynamics model achieved an average maximum position prediction error of only \textit{0.43\,cm} across all prediction windows. In contrast, the average maximum deviation between the commanded and executed end-effector trajectories reached \textit{4.57\,cm}.
Such execution mismatch is sufficiently large to severely degrade long-horizon predictive planning and can become catastrophic for fast dynamic manipulation, highlighting the importance of execution-aware prediction.

A test-time prediction example is shown in Fig.~\ref{Execution Dynamics Model}. 
The blue curve denotes the commanded trajectory, while the black curve corresponds to the measured EE trajectory of the Franka robot. 
A clear execution delay can be observed between the commanded and executed motions. 
The red curve represents the trajectory predicted by the execution dynamics model, which closely matches the measured EE motion throughout the prediction horizon.

\subsubsection{Trajectory Tracking}

\begin{table}[t]
\centering
\caption{BM2PC and controller parameters used in the real-world experiments.}
\label{tab:bm2pc_real_hyperparameters}
\small
\setlength{\tabcolsep}{3pt}
\renewcommand{\arraystretch}{1.08}
\begin{tabular}{lcc}
\toprule
Parameter & Tracking & Obstacle Avoid. \\
\midrule
Prediction horizon $H$ & 50 & 50 \\
Number of modes $n$ & 1 & 3 \\
Samples / iteration $N$ & 81 & 81 \\
Optimization iterations $I$ & 1 & 2 \\
Sampling std. $\sigma$ & 1.8 & 1.8 \\
Sampling covariance $\Sigma$ & $\sigma^2 I$ & $\sigma^2 I$ \\
Cost weight $\lambda^{-1}$ & 3000 & 300 \\
Diversity weight $\beta$ & 200 & 200 \\
Retained candidates $K$ & 81 & 81 \\
Filtering threshold & $10^5$ & $10^5$ \\
Acceleration limits (m/s$^2$) & $\pm5$ & $\pm5$ \\
MPC frequency (Hz) & 66.67 & 66.67 \\
Low-level frequency (Hz) & 1000 & 1000 \\
\bottomrule
\end{tabular}
\end{table}

The BM2PC and controller parameters used in the real-world experiments are summarized in Table~\ref{tab:bm2pc_real_hyperparameters}. 
For trajectory tracking, BM2PC uses a single control mode ($n=1$), a prediction horizon of $H=50$, $N=81$ samples per optimization iteration, and one optimization iteration per control step. 
The same parameter setting is used for all three tracking trajectories without trajectory-specific retuning.

Due to the limited reachable workspace of the robotic arm, the target trajectories are scaled down for the real-world experiments. Additionally, the \textit{Lemniscate} trajectory is repeated for three consecutive cycles to evaluate long-term tracking stability. The parameters of the three trajectories are summarized in Table~\ref{tab:real_traj_tracking}.

\begin{figure}[t]
    \centering

    \includegraphics[width=1.0\linewidth]{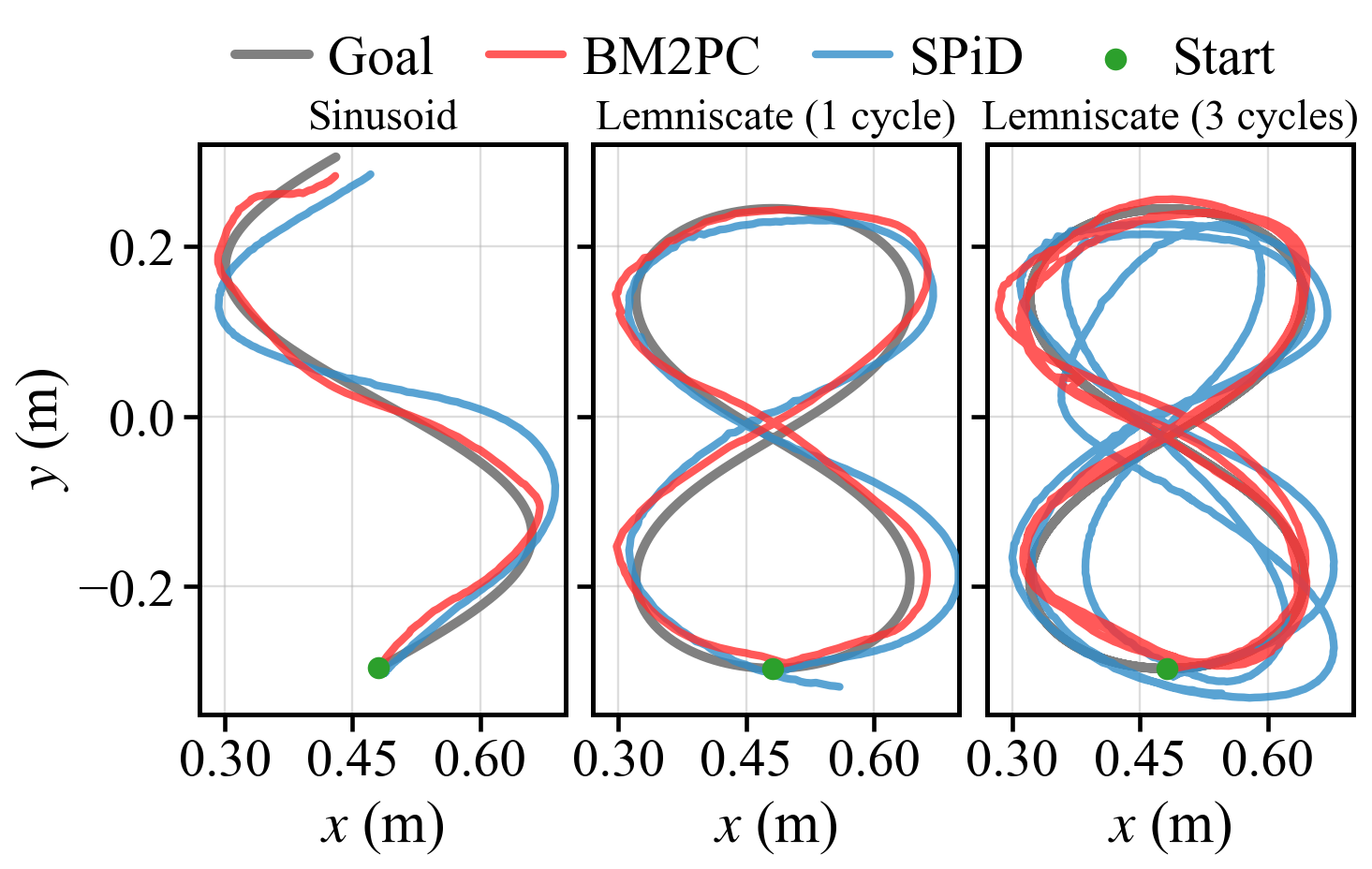}

    \vspace{-0.0cm}

    \begin{flushright}
        \includegraphics[width=0.92\linewidth]{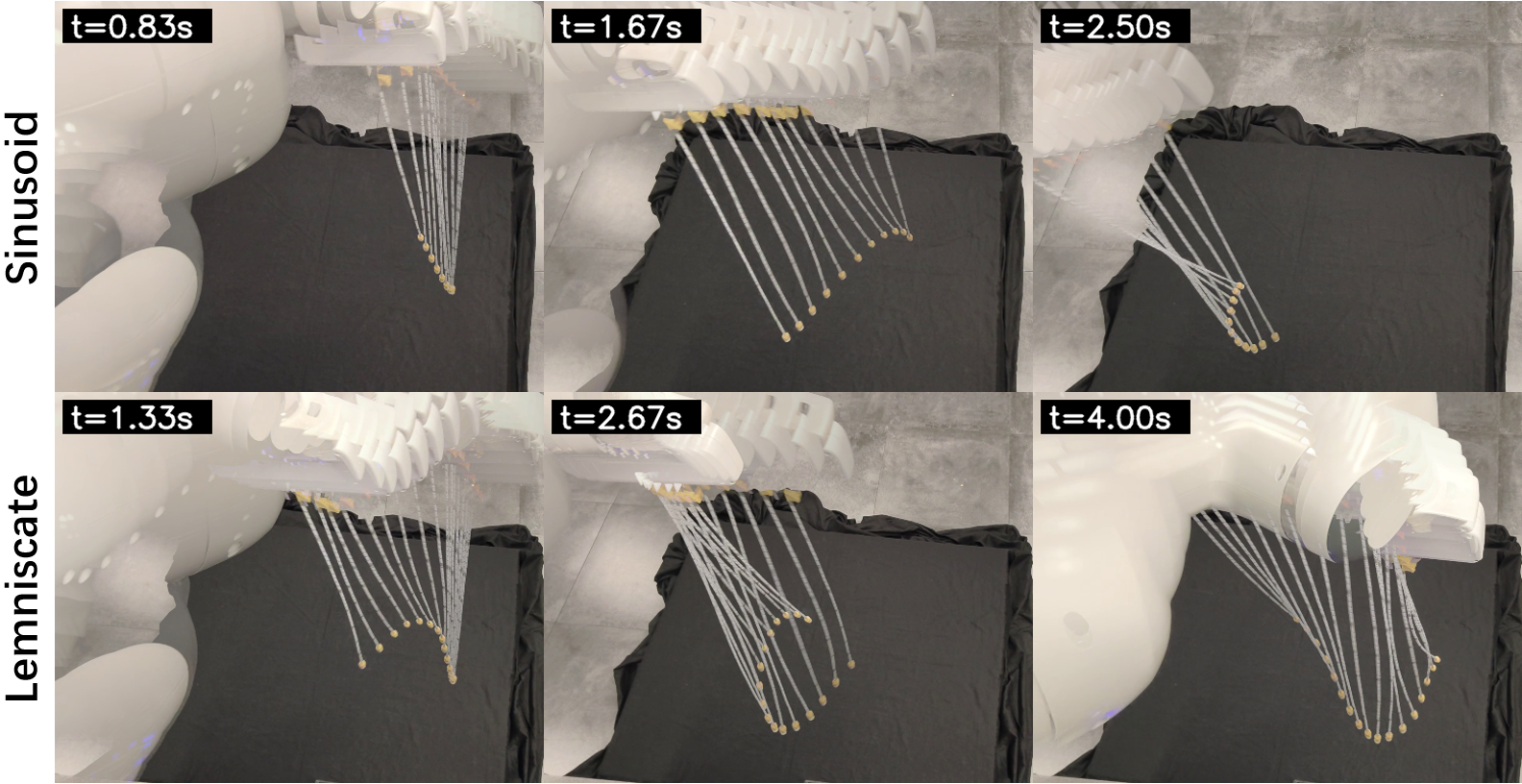}
    \end{flushright}

    \caption{
    \textbf{Real-world trajectory tracking results.}
    Top: comparison of the desired trajectory and the tracked trajectories generated by BM2PC and SPiD for three trajectories.
    Bottom: time-lapse snapshots of the real robotic system during the sinusoidal and lemniscate tracking experiments by BM2PC. 
    }
    \label{fig:real-tracking}
\end{figure}

The trajectory tracking results are shown in Fig.~\ref{fig:real-tracking}. It can be observed that the proposed method achieves higher tracking accuracy across all trajectories. Moreover, during the three-cycle lemniscate tracking task, BM2PC maintains consistent tracking performance over the entire trajectory, whereas the tracking error of SPiD gradually accumulates over time. Table~\ref{tab:real_traj_tracking} further provides a quantitative comparison in terms of RMSE. 
The results reinforce the conclusions of the simulation results (Sec.~\ref{sec:exp-sim-traj}) that the model-based approach is superior to the learned baseline in terms of generalization, including sim-to-real transfer. 

We further conducted an ablation study on the three-cycle lemniscate task by removing the execution dynamics model. Without execution-aware prediction, the tracking RMSE increased dramatically to \textit{23.68\,cm}, and the controller failed to reliably track the target trajectory. These results highlight the importance of execution-aware prediction for long-horizon dynamic manipulation under real-world execution mismatch.

\begin{table}[t]
\centering
\caption{Real-world trajectory tracking task specifications and tracking RMSE comparison.}
\label{tab:real_traj_tracking}


\begin{tabular}{@{}lccc@{}}
\toprule
& \textbf{Sinusoid} & \textbf{Lemn. ($\times$1)} & \textbf{Lemn. ($\times$3)}\\
\midrule
Duration (s)       & 2.500  & 4.000  & 10.800 \\
Length (m)         & 0.932 & 1.806 & 5.420 \\
Max Velocity (m/s) & 0.593 & 0.699 & 0.680 \\
\midrule
SPiD RMSE (cm)     & 4.300 & 3.451 & 4.414 \\
BM2PC RMSE (cm)     & \textbf{4.175} & \textbf{2.701} & \textbf{2.474} \\
\bottomrule
\end{tabular}

\end{table}

\begin{figure*}[h]
      \centering
      \includegraphics[width=0.89\linewidth]{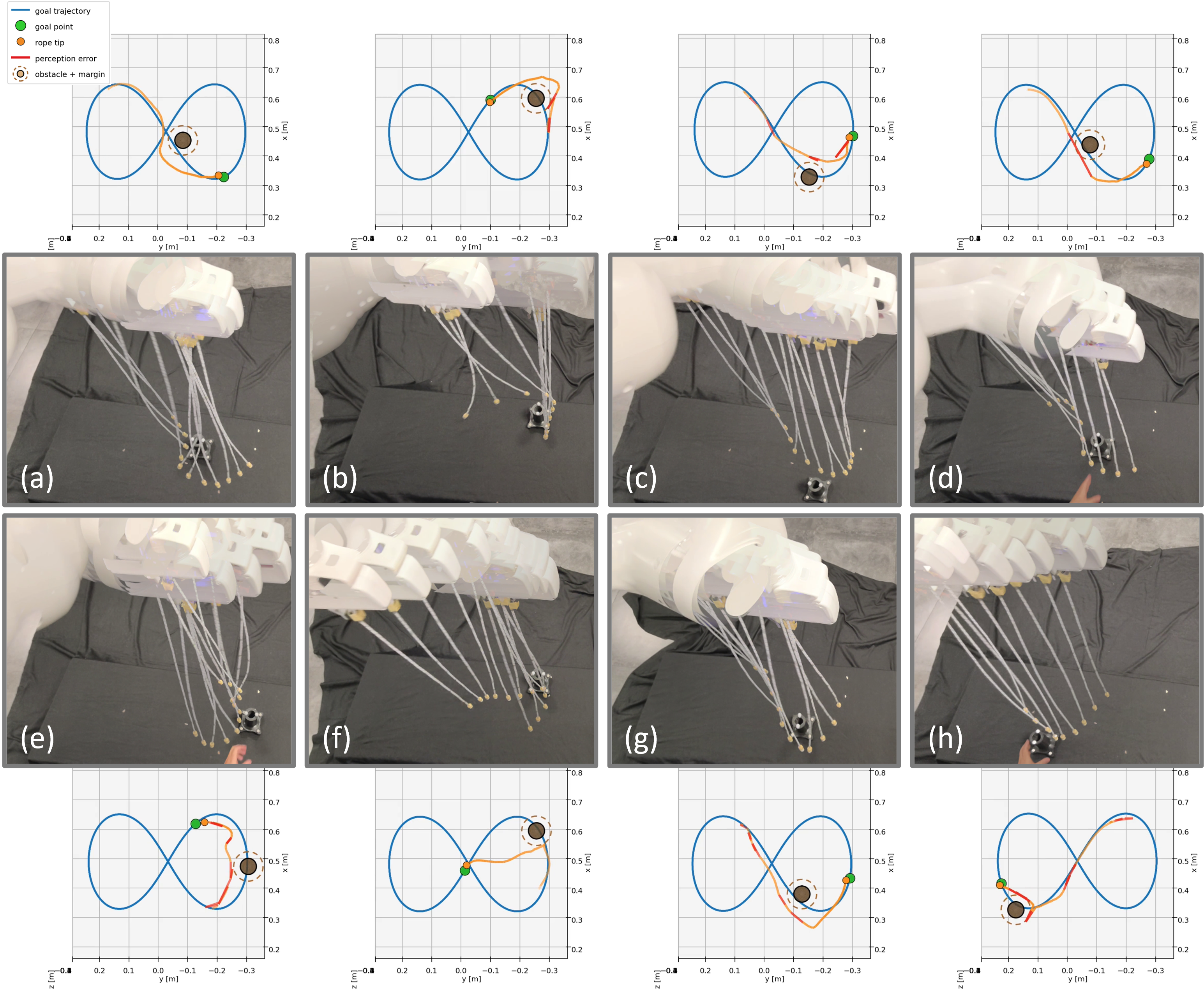}
      \caption{\textbf{Real-world obstacle avoidance results.} The rope tracks a Lemniscate ($\times 3$) trajectory while a movable obstacle is randomly repositioned along the path to evaluate dynamic obstacle avoidance performance. The blue curve denotes the target trajectory, the orange curve represents the executed trajectory, and the red segments indicate perception failures caused by occlusions, environmental noise, or network instability. Time-lapse snapshots (a)–(h) illustrate the avoidance process under different obstacle configurations. Despite severe perception degradation, including frozen measurements and abrupt trajectory spikes, the controller successfully avoids the obstacle and rapidly returns the rope endpoint to the desired trajectory.}
      \label{real obs}
\end{figure*}

\subsubsection{Obstacle Avoidance}

For obstacle avoidance, BM2PC maintains three control modes ($n=3$) to represent distinct avoidance behaviors and uses two optimization iterations per control step. 
Compared with trajectory tracking, the cost weight is reduced to $\lambda^{-1}=300$, while the remaining sampling and controller settings are unchanged, as summarized in Table~\ref{tab:bm2pc_real_hyperparameters}.

In the obstacle-avoidance experiments, the rope was commanded to track a \textit{Lemniscate~($\times 3$)} trajectory while a movable obstacle was placed at and randomly repositioned along the path to evaluate dynamic obstacle-avoidance performance. \highlight{We conducted three trials with different obstacle placements, all of which were completed without collision. BM2PC achieved a mean trajectory-tracking RMSE of $4.51 \pm 1.11$\,cm and a mean peak tracking deviation of $15.67 \pm 2.16$\,cm. For avoidance events in which the tracking error exceeded 5\,cm, the mean recovery time was $0.78 \pm 0.61$\,s, with a maximum of 1.72\,s. Here, recovery time is measured from the peak tracking error until the error remains below 5\,cm for at least 0.3\,s.} Time-lapse snapshots are shown in Fig.~\ref{real obs}, and the complete avoidance process is provided in the accompanying video.

However, during obstacle avoidance, the primary challenge arises not from control but from the perception system. The red lines in Fig.~\ref{real obs} correspond to position jumps due to perception failures, which account for approximately 4.96\% of all measurements during the experiment. In our setup, more than half of the camera viewpoints are occluded by the table, robotic arm, and obstacles, which can lead to temporary loss of the markers and consequently unavailable rope state measurements. 
The occlusions increase particularly around the obstacles.
In these cases, the estimated measurements remain nearly unchanged for several consecutive frames, appearing “frozen’’ at outdated positions. Once the markers are reliably detected again, the estimated positions abruptly jump to the updated locations. 
In addition, due to environmental noise and network instability, the measurements may occasionally jump to incorrect positions before rapidly returning to the true trajectory, resulting in several-frame spikes.
Despite these severe perception degradations, the controller remains capable of successfully accomplishing the task, validating its robustness.

\section{CONCLUSIONS}

We introduced BM2PC, which formulates multimodal MPC as Bayesian set optimization and enables efficient maintenance of multiple low-cost, behaviorally distinct control hypotheses.
We evaluated the proposed method on dynamic DLO manipulation tasks involving trajectory tracking of complex, hand-drawn target trajectories, including non-smooth motions that are difficult to predict and execute. 
The results demonstrated that our model-based approach consistently outperforms the learning-based baseline, demonstrating stronger performance on previously unseen tracking tasks. This advantage became even more pronounced in real-world experiments, where sim-to-real discrepancies further increased task diversity beyond the distribution encountered during training.

We further compared BM2PC against state-of-the-art multimodal MPC methods on obstacle-avoidance tasks, where maintaining multiple candidate solutions is essential. 
The results showed that BM2PC achieves comparable solution quality while requiring substantially less computation on the same hardware, highlighting the effectiveness of the proposed approximation. 
The disturbance rejection experiments further demonstrate the robustness of predictive control under large external perturbations, while the obstacle avoidance results highlight the benefits of maintaining multiple control strategies under changing constraints.

Finally, successful real-world deployment was enabled by the proposed execution dynamics model, which compensates for command-execution mismatch, thereby providing the planner with a more accurate prediction of the executed robot motion.
Together, these contributions establish BM2PC as an efficient, robust, and practical framework for real-time multimodal control of deformable linear objects, bridging the gap between computationally demanding multimodal planning and real-world robotic manipulation.







\balance
\bibliographystyle{IEEEtran}
\bibliography{ref}   

\end{document}